%% file: iclr2025_conference.tex
\documentclass{article} 
\usepackage{iclr2025_conference,times}

\input{math_commands.tex}

\usepackage{hyperref}
\usepackage{url}

\usepackage[pdftex]{graphicx}
\usepackage{subcaption}
\usepackage{booktabs}
\usepackage{pifont}
\usepackage{kotex}
\usepackage{multirow}
\usepackage{float}

\title{Adapters for Altering LLM Vocabularies: What Languages Benefit the Most?}

\stepcounter{footnote}\stepcounter{footnote}
\author{HyoJung Han\thanks{Work done at Microsoft.}\\
University of Maryland\\
\texttt{\small{hjhan@cs.umd.edu}}
\And
Akiko I. Eriguchi\\
Microsoft\\
\texttt{\small{akikoe@microsoft.com}}
\And
Haoran Xu\\
Microsoft\\
\texttt{\small{haoranxu@microsoft.com}}
\And
Hieu Hoang\\
Microsoft\\
\texttt{\small{hihoan@microsoft.com}}
\And
Marine Carpuat\\
University of Maryland\\
\texttt{\small{marine@cs.umd.edu}}
\And
Huda Khayrallah\footnotemark[3]\\
Amazon\\
\texttt{\small{hudakh@amazon.com}}
}

\newcommand{\adt}[0]{VocADT} 

\newif\ifcomment\commenttrue
\ifcomment
    \newcommand{\customcmt}[3]{\textcolor{#1}{[#2: #3]}}
\else
    \newcommand{\customcmt}[3]{}
\fi

\iclrfinalcopy 
\begin{document}

\maketitle

\begin{abstract}
Vocabulary adaptation, which integrates new vocabulary into pre-trained language models, enables expansion to new languages and mitigates token over-fragmentation.
However, existing approaches are limited by their reliance on heuristics or external embeddings.
We propose \adt{}, a novel method for vocabulary adaptation using adapter modules that are trained to learn the optimal linear combination of existing embeddings while keeping the model's weights fixed. \adt{} offers a flexible and scalable solution without depending on external resources or language constraints. 
Across 11 languages---with diverse scripts, resource availability, and fragmentation---we demonstrate that \adt{} outperforms the original Mistral model \citep{mistral7b} and other baselines across various multilingual tasks %
including natural language understanding and machine translation.
We find that Latin-script languages and highly fragmented languages benefit the most from vocabulary adaptation.
We further fine-tune the adapted model on the generative task of machine translation and find that vocabulary adaptation is still beneficial after fine-tuning and that \adt{} is the most effective%
.\footnote{Project page: \url{https://github.com/h-j-han/VocADT}. Models at \href{https://huggingface.co/collections/h-j-han/vocadt-67084ac852855267504fd0c6}{\texttt{Huggingface Hub}}
}
\end{abstract}

\section{Introduction}
\label{sec:intro}

Vocabulary adaptation (or transfer)---a process of modifying a pre-trained language model (LM) to use a new vocabulary---offers several key advantages. 
First, it enables the introduction of new languages into a model, increasing flexibility in handling linguistic diversity and improving downstream performance in target languages \citep{wang-etal-2020-extending, gogoulou-etal-2022-cross, downey-etal-2023-embedding}.  %
Second, it reduces over-fragmentation, where words are excessively split by the tokenizer, slowing down generation\footnote{Standard transformer decoding is quadratic in sequence length, so length increases can be catastrophic.} and impairing performance in certain languages \citep{do-all-lang-cost-same23, token-unfairness23, yamaguchi2024empirical}. 
These benefits have led to the development of numerous vocabulary adaptation approaches
that initialize the new embeddings of new vocabulary with various methods based on heuristics \citep{vipi22, fvt22, downey-etal-2023-embedding}, external resources \citep{ramen20, focus23, ofa24}, or a separate hypernetwork that generates it \citep{zett24}.
They generally generate new embeddings using original embeddings, 
optionally followed by continued training to finalize the adaptation \citep{wechsel22, clp23, focus23, ofa24}.

However, existing vocabulary adaptation approaches face several limitations. 
Those that rely on heuristics \citep{fvt22, downey-etal-2023-embedding}, which use predefined rules to initialize new embeddings from existing ones rather than learning from data, often lack adaptability, where the new embeddings are not fully integrated into the original model and require an additional training phase of full-weight updates to fully adapt to the new vocabulary.
Also, those that depend on external embeddings or networks \citep{ramen20, focus23, ofa24}, increase complexity and limit scalability. 
Furthermore, some approaches focus solely on language-specific cases or may have restrictions on the number of languages in the implementation when configuring the vocabulary \citep{focus23, zett24}.

Additionally, we still know little about the impact of vocabulary adaptation across diverse linguistic and task settings.
Most prior work investigates few languages, which is insufficient for identifying patterns
\citep{ramen20, clp23,tiktotok23, yamaguchi2024empirical}, while studies that consider a broader range of languages only report averages without detailed analysis
\citep{ofa24, cw2v24}.
Furthermore, the impact of vocabulary adaptation on cross-lingual and generative tasks like machine translation (MT) is understudied, even though they represent crucial application areas for porting models to new languages. Many adaptation methods \citep{chung-etal-2020-improving,fvt22, ofa24} have been evaluated instead on non-cross-lingual and discriminative tasks such as commonsense reasoning, natural language inference (NLI), or question answering (QA)---which are typically classification tasks. 

\begin{figure}[t]
\begin{center}
    \centering
    \subcaptionbox{Overview of the vocabulary adaptation and training. %
    \label{fig:train}}
    [.54\linewidth]{\includegraphics[scale=0.99]{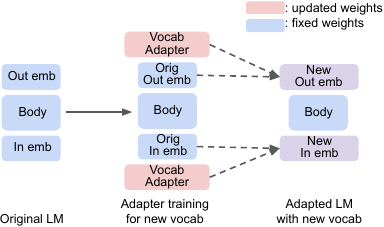}}
    \subcaptionbox{Initialization of vocabulary adapter.
    \label{fig:initialization}}
    [.45\linewidth]{\includegraphics[scale=0.65]{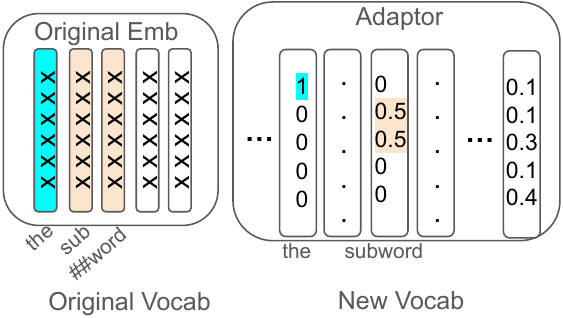}}
\end{center}
\caption{
Overview of our vocabulary adaptation with adapter (\adt{}) and the initialization of adapter. 
The vocabulary adapter modules are trained to adapt new vocabulary with existing embeddings while keeping the original model fixed. 
We initialize entries of the adapter for overlapping tokens and tokens whose partitions are in the original vocabulary. 
Once trained, the adapters and original embeddings are merged to form the new embeddings.
}
    \label{fig:overview}
\end{figure}

We propose \adt{}, a novel solution for vocabulary adaptation using adapters, designed to address key challenges in existing approaches (Figure~\ref{fig:overview}). 
We introduce a vocabulary adapter module, a learnable matrix between the new vocabulary and the original embeddings of a language model.
The module gradually adapts to new vocabularies through training while keeping all weights of the original model fixed, allowing the module to learn the best combination of the original embeddings without relying on heuristics, external embeddings, or dictionaries.
This learned adaptation approach offers better adaptability of new embeddings to the original language model with only its embeddings replaced
and more flexibility in the number of languages while removing the necessity of external pre-trained resources.
At the end of training, the adapter is
merged with the original embeddings to create a new embedding matrix.

In addition to our novel method, we empirically address the following key questions to understand the effectiveness and behavior of vocabulary adaptation: 
(1) Which languages benefit most from vocabulary adaptation?
(2) What are the best strategies for creating new vocabularies? Also, is script consistency necessary? 
(3) How does vocabulary adaptation impact machine translation?
We emphasize this task as it is a critical task for multilingual models that involves cross-lingual and generative capabilities, which are often more complex than classification or monolingual tasks.

We demonstrate the effectiveness of our adaptation method on various NLP tasks spanning Natural Language Understanding and MT.
Results show that our approach consistently surpasses the original Mistral model in most cases, both after the adaptation phase and following phase of full-weight training.
Additionally, our method outperforms or matches other strong vocabulary adaptation baselines. %
Our findings indicate that Latin-script languages and those with severe fragmentation benefit the most from vocabulary adaptation. %
Finally, while all vocabulary adaptation methods continue to be effective for machine translation after fine-tuning, \adt{} shows the best results among them.
Our main contributions are summarized as follows:
\begin{itemize}
    \item We propose \adt{}, a simple and effective solution for vocabulary adaptation using adapters, that addresses key limitations in prior work such as reliance on external embedding or language constraints.    
    \item We conduct experiments that cover a wide range of languages and scripts, finding that languages with Latin scripts or severe fragmentation benefit the most and that having a consistent grouping of scripts for multilingual vocabulary is helpful.
    \item  Our approach consistently outperforms the original language model and is more effective than, or on par with, strong vocabulary adaptation baselines after the adaptation phase across various tasks and after the following full-weight fine-tuning on MT.
\end{itemize}

\section{Background and Motivation}
\label{sec:background}

\paragraph{Approaches to Vocabulary Adaptation} Prior work focuses on initialization strategies for the new vocabulary embeddings, before continuing training with unlabeled target language text using the original self-supervised pretraining objective. For instance, FOCUS~\citep{focus23} initializes embeddings as a weighted combination of overlapping tokens using external embeddings for non-overlapping ones, while copying the embeddings of overlapping tokens. OFA~\citep{ofa24} also relies on external word vectors to initialize embeddings for non-shared new tokens, using a weighted average of original tokens based on semantic similarity. This strategy often requires external resources such as auxiliary embeddings \citep{focus23, ofa24, clp23} or bilingual dictionaries \citep{cw2v24, wechsel22}.

After initialization, language adaptive pretraining \citep[LAPT;][]{chau-etal-2020-parsing} usually updates all model weights \citep{ramen20, ofa24, focus23, clp23}, except \citet{yamaguchi2024empirical} which use LoRA \citep{hu2022lora}. \citet{downey-etal-2023-embedding} show full-weight updates outperform embedding-only training, which is insufficient for multilingual transfer. 

Other vocabulary adaptation strategies introduce architecture-specific changes to the model, such as MAD-X~\citep{madx20}, which incorporates various adapters into Transformer models, and thus additional computation costs. There are few alternatives to these resource-intensive approaches. A notable exception is ZeTT~\citep{zett24}, which trains a hypernetwork that generates embeddings for the new vocabulary, allowing immediate zero-shot use by only replacing embeddings without further model training. It can be extended to multilingual hypernetworks by appending a learnable language-specific embedding.

\paragraph{Linear Combination of Embeddings} 
Most vocabulary transfer methods combine the existing embeddings to 
generate new ones.
A popular approach is to use a weighted average of the original embeddings (\textbf{bolded} in  Appendix Table~\ref{table:appndx_related_works}).
For example, \citet{fvt22} and \citet{vipi22} compute the new embeddings by simply averaging the embeddings of subword tokens, while \citet{ramen20}, \citet{wechsel22}, OFA, and FOCUS utilize external resources to determine the weights to initialize the new embeddings with a weighted average of the original embeddings. 
\citet{cw2v24} established theoretically that initializing within the convex hull of existing embeddings—e.g., using a weighted average of source embeddings—is a good initialization.

Our motivation stems from the question: rather than deciding how to combine existing embedding vectors heuristically, why not learn this process to create new embedding vectors?
Relying on heuristics may lack adaptability that typically requires an additional training phase of full-weight updates to fully adapt to the new vocabulary. 
Building upon prior works, we propose to learn linear combinations with vocabulary adapters.

\paragraph{Empirical Evaluations} Many language adaptation experiments have been conducted using new language-specific monolingual vocabularies \citep{zett24, focus23, madx20, wechsel22, yamaguchi2024empirical}, as well as English-only but domain-specific vocabularies \citep{fvt22, vipi22}. 
In contrast, \citet{ofa24} and \citet{cw2v24} use a single unified multilingual vocabulary covering at least 369 languages and four languages, respectively. 

\citet{downey-etal-2023-embedding} conducted experiments with both monolingual and multilingual vocabularies across eight languages and additional vocabulary from the Uralic family. While their findings indicated that multilingual adaptation in the Uralic family followed overall trends,  it remains unclear whether vocabulary adaptation benefits languages in different script groups. Overall, empirical evidence is still lacking to guide practical decisions for grouping languages in multilingual models. 

Furthermore, most studies exclusively evaluate models on non-cross-lingual and non-generative tasks, such as binary or multi-class classification, sequence labeling (e.g., part-of-speech tagging), or answer span prediction. \citet{cw2v24} and \citet{yamaguchi2024empirical} include monolingual generative tasks like summarization. As a result, the impact of vocabulary adaption on cross-lingual generation tasks such as MT remains understudied, even though this is a crucial application area. \footnote{
\citet{cw2v24} report MT results, but they do not release the models or per-language performance metrics, making direct comparisons difficult.
} 

To address these gaps, this work introduces a strategy to adapt LLM to new vocabularies, and experiments designed to measure its impact across diverse linguistic and task settings. We compare empirically against FOCUS and OFA, representing more resource-intensive initialization approaches, and ZeTT, representing a more parsimonious approach that has only been tested on limited tasks so far. Summaries of prior work can be found in Appendix Table~\ref{table:appndx_related_works}.

\section{\adt{}: Multilingual Vocabulary Adaptation with Adapters}
\label{sec:method_adt}
In this section, we outline our approach to multilingual vocabulary adaptation using adapter modules (\adt{}). 
We detail the architecture of the adapter module (\textsection \ref{sec:method_adt_arch}) and the initialization process (\textsection \ref{sec:method_adt_init}). 
Additionally, we introduce an additional loss for handling overlapping tokens between the new and original vocabularies (\textsection \ref{sec:method_adt_auxloss}).
Finally, we further fine-tune the vocabulary adapted model for downstream task (\textsection \ref{sec:method_alma}).

\subsection{Vocabulary Adapter Module}
\label{sec:method_adt_arch}
We introduce the vocabulary adapter modules to find parameters of new embeddings that can replace the original embedding without changing the non-embedding part of the original model (Figure~\ref{fig:train}).
For simplicity, we refer to both input and output embeddings (or the LM head) collectively as embeddings.
Let $V^o$ and $V^n$ be the $\bm{o}$riginal and $\bm{n}$ew vocabulary, respectively, and let $\mathcal{T}^x: w \rightarrow (t_1, t_2, \dots, t_k)$ be a tokenizer associated with a vocabulary $V^x$ where $t_j \in V^x, \forall j = 1, \dots, k$.
We put vocabulary adapter modules $ \mA \in \R^{|V^n|\times |V^o|}$ between the new vocabulary $V^n$ and the original embedding $\mE^o \in \R^{|V^o|\times h}$ where $h$ is an embedding dimension, in a manner similar to bottleneck adapters \citep{pmlr-v97-houlsby19a}.
We train the adapters with the standard language modeling loss $\mathcal{L}^{lm}$, where we freeze the original weights and only update the adapters.
This may be analogous to finding the new embedding vector for a token with the weighted combination of original embedding vectors \citep{downey-etal-2023-embedding,focus23,ofa24}. Unlike similar works, our approach learns the weights for embedding combination.
After training the adapters, we get new embeddings $\mE^n \in \R^{|V^n|\times h}$ by merging the original embeddings and adapters to $\mE^n = \mA\mE^o$, which results in a language model with the same architecture as the original one but with a different vocabulary size.

\subsection{Initializing Adapter}
\label{sec:method_adt_init}
Effective initialization of the new embedding is crucial in adapting to a new vocabulary, as fully random initialization is widely recognized for leading to poor performance \citep{wechsel22, yamaguchi2024empirical}. 
In our case, random initialization of the adapter $\mA^{0}$ is equivalent to random initialization of $\mE^n$, making proper initialization of \(\mA^{0}\) equally important.
We suggest a simple initialization scheme for the vocabulary adapter, illustrated in Figure~\ref{fig:initialization}.

First, we follow the common methods of copying the original embeddings of overlapping tokens by setting a one-hot vector in the adapter. Let $\mathcal{I}^x: V^x \rightarrow \mathbb{Z}$ be the mapping function of a token to an index in a vocabulary $V^x$ and let $i = \mathcal{I}^n(w)$ be the index of a token $w$ in $V^n$. The row of the adapter $\mA^{0}_{i}$ corresponding to the overlapping tokens %
$w \in V^o \cap V^n$ is set as follows:
\begin{equation}
\mA^{0}_{i, \mathcal{I}^o(w)} = 1, \quad \mA_{i, j} = 0 \quad \forall j \neq \mathcal{I}^o(w), \quad \text{where} \ w \in V^o \cap V^n.
\label{eq:init_ovl_whole}
\end{equation}

Inspired by \citet{fvt22}, we then initialize the row of a token $w$ in $\mA^0$, whose partitioned tokens by the original tokenizer $\mathcal{T}^o$ are subset of the original vocabulary, $\mathcal{T}^o(w) = \{t_1, \dots, t_m\} \subset V^o, m > 1$, with normalized multi-hot vector as below. This corresponds to directly initializing new embedding %
with the average of the original embeddings associated with the tokens produced by $\mathcal{T}^o$.
\begin{equation}
\mA^{0}_{i, j} = \begin{cases}
\frac{1}{m} & \text{if } j \in \{\mathcal{I}^o(t_1), \dots, \mathcal{I}^o(t_m)\} \\
0 & \text{otherwise}
\end{cases}  \text{where} 
\begin{array}{l}
w \in V^n \backslash (V^o \cap V^n) \quad \text{and}\\ w \in S=\{ w \mid \mathcal{T}^o(w) = \{t_{1:m}\} \subset V^o\}.
\end{array}
\label{eq:init_ovl_sub}
\end{equation}

For a token that does not fall into the first two cases above (i.e. a non-overlapping token and its partitions by $\mathcal{T}^o$ are not in $V^o$), we randomly initialize a row vector of the adapter with the uniform distribution whose sum of each element is one as follows:
\begin{equation}
\mA^{0}_{i} = \frac{\mathbf{u}}{\sum_{j=1}^{|V^o|} u_j}, \quad u_j \sim \text{Uniform}(0, 1), \ j= 1, \dots, |V^o|\quad  \text{where} \ w \in V^n \backslash (V^o \cap V^n) \backslash S. %
\label{eq:init_non_ovl}
\end{equation}

\subsection{Auxiliary Loss}
\label{sec:method_adt_auxloss}
As training progresses, the adapter entries of overlapped tokens tend to diverge from their initial states. 
This divergence can be undesirable because the original embeddings are already well-integrated into the language model, and our goal is more focused on adjusting the embeddings of the newly introduced vocabulary items. 
Following \citet{zett24}, we experiment with an additional loss term that encourages the adapter entries for overlapping words to remain close to their initial values, formulated as follows:
\begin{equation}
\mathcal{L}^{aux} = \frac{1}{|V^o \cap V^n|} \sum_{w \in |V^o \cap V^n|} || \mA_{\mathcal{I}^n(w)}  - \mA^{0}_{\mathcal{I}^n(w)} ||_2 .
\label{eq:loss_ovl}
\end{equation}

The final loss for the adapter training is the combination of the standard language loss and additional loss with the weighing factor of $\alpha$, $\mathcal{L}^{tot} = \mathcal{L}^{lm} + \alpha\mathcal{L}^{aux}$.

\subsection{Further Fine-tuning For Downstream Task}
\label{sec:method_alma}
To understand the impact of 
vocabulary adaptation  after task-specific fine-tuning,
we follow the full ALMA~\citep{alma24} training 
on all model parameters for the cross-lingual generation task of machine translation after our  \adt{} on just the embeddings. 
ALMA training begins with fine-tuning on monolingual data, followed by further weight optimization on small curated parallel data.

\section{Which Languages Benefit the Most from Vocabulary Adaptation?}
\label{sec:method_vocab}
We aim to understand ``When and how should we perform vocabulary adaptation?''.
More specifically, we seek insight into which languages might benefit the most from vocabulary adaptation in terms of improving overall performance or mitigating over-fragmentation.\footnote{
The non-English languages that we cover are all highly fragmented by common LLMs, 
and their fragmentation is similarly improved by our method. Therefore, our analysis focuses on performance. }

To this end, we design experiments to cover 10 non-English languages along with English, listed in Table~\ref{table:langs}, with a variety of scripts and language families.
These languages are broadly categorized into three groups:
(1) \textit{Latin} group of Swahili, Indonesian, Estonian, and Haitian, which are low- to mid-resource languages and all use Latin script;
(2) \textit{Mixed} group including Korean, Greek, Russian, and Bulgarian, which utilize a mixture of scripts;
(3)  \textit{Cyrillic} group for languages with that scrip.\footnote{In Section~\ref{sec:result_all_lma} and Appendix~\ref{sec:result_all}, we experiment with \textit{All} group including all languages mentioned here.}

We test individual language adaptation with language-specific vocabularies. 
We also adapt several multilingual vocabularies that include English and four non-English languages in a single shared vocabulary, with each group corresponding to one of the previously mentioned groups.
This is to identify a language grouping strategy---whether to mix languages with different scripts or grouping languages with consistent scripts.

\input{tables/langs}

\section{Experiment Design}
\label{sec:exp}

\subsection{Baselines and Modeling}
\label{sec:exp_model}

We use Mistral-7B~\citep{mistral7b} as our language model, along with its original vocabulary, which consists of 32k tokens ($|V^o|=32k$).
As baselines, we evaluate three state-of-the-art methods for vocabulary adaptation, ZeTT~\citep{zett24}, FOCUS~\citep{focus23}, and OFA~\citep{ofa24}. %
For ZeTT and FOCUS, we experiment with language-specific vocabularies (ZeTT-mono, FOCUS-mono) as their implementations require specifying the language to adapt the vocabulary.
This results in separate adaptations per language, which could be hard to scale with larger language coverage.\footnote{
ZeTT does not support Ukrainian and Kazakh, therefore we primarily compare and average the results for 9  languages covered by both methods. 
See Appendix~\ref{sec:apdx_result_number_langwise} for results for Ukrainian and Kazakh.}
For \adt{} and OFA methods (\adt{}-multi, OFA-multi), we experiment with multilingual vocabularies of five languages including English and four non-English languages, where we define three distinct language groups such as $\{$\texttt{en}, \texttt{sw}, \texttt{id}, \texttt{et}, \texttt{ht}$\}$ ( \textit{Latin} group), $\{$\texttt{en}, \texttt{ko}, \texttt{el}, \texttt{ru}, \texttt{bg}$\}$ (\textit{Mixed} group), and $\{$\texttt{en}, \texttt{ru}, \texttt{bg}, \texttt{uk}, \texttt{kk}$\}$ (\textit{Cyrillic} group).

\subsection{Training \adt{}}
\label{sec:exp_train}

\paragraph{Vocabulary.} 
We train SentencePiece~\citep{sentencepiece18} tokenizers on either language-specific corpora or a combined corpus, with a maximum of 2 million tokens per language, and create new vocabularies with a size of 50k for all cases including mono/multilingual vocabularies ($|V^n|=50k$). 
Newly created vocabularies for each language group are shared across baselines. %

\paragraph{Adapter Training.} 
In the adapter training phase, we train only the adapters, while fixing all parameters of the original model. %
The input and output adapters are separate modules, as preliminary results showed that sharing an adapter for the input and output sides performs worse.
We train 0.5B monolingual tokens per language, totaling 2.5B mixed by 5 languages (English + 4 non-English from each corresponding group), and report test numbers from it.
We use ``clean'' documents from the corpus of MADLAD-400~\citep{madlad23}. %
We set the weighing factor of auxiliary loss $\alpha$ with 0.1 for non-Latin groups %
and 0 for the \textit{Latin} group unless otherwise specified.
This is based on the empirical results in Appendix~\ref{sec:result_adt_ovl} %
that maintaining the embeddings of overlapping tokens close to the original status during the adaptation is effective only for non-Latin script languages and counter-effective for Latin languages.
More details regarding the training are in Appendix~\ref{sec:apdx_training_details}.

\subsection{Full-Weight Fine-tuning}
After the adaptation phase, we follow the fine-tuning recipe of ALMA~\citep{alma24} that consists of full-weight training with monolingual corpus and a small amount of high-quality parallel corpus to enhance MT performance (\textsection\ref{sec:method_alma}).
We also include Mistral in this phase of training.
1) In monolingual fine-tuning, we use MADLAD-400. %
For adapted ZeTT and FOCUS models (prior work), we fine-tune each separate model with non-English language-specific vocabulary except for \texttt{uk} and \texttt{kk} for ZeTT (again, due to unsupported languages in ZeTT) using a total of 2B tokens combining English and the corresponding non-English. %
For Mistral and the adapted \adt{} and OFA, we fine-tune separate models for all three non-English groups (\textit{Latin}, \textit{Mixed}, \textit{Cyrillic}) plus English using a corpus of 5B monolingual tokens containing 5 languages.
2) In the next parallel training, we sample 15k bitext from NLLB dataset \citep{schwenk-etal-2021-ccmatrix, heffernan-etal-2022-bitext, flores22}\footnote{\url{https://huggingface.co/datasets/allenai/nllb}} for each English and non-English training pairs with top LASER3 scores \citep{laser19}.
The parallel training is done for one epoch, and we report test set numbers with the best model of the validation set.
All the models are fine-tuned and tested with both directions of \texttt{en-xx} and \texttt{xx-en} within a single model, meaning there are no separate models for opposite translation directions.
We follow the prompting strategy of \citet{alma24}. %

\begin{figure}[t]
\begin{center}
    \includegraphics[width=1\linewidth]{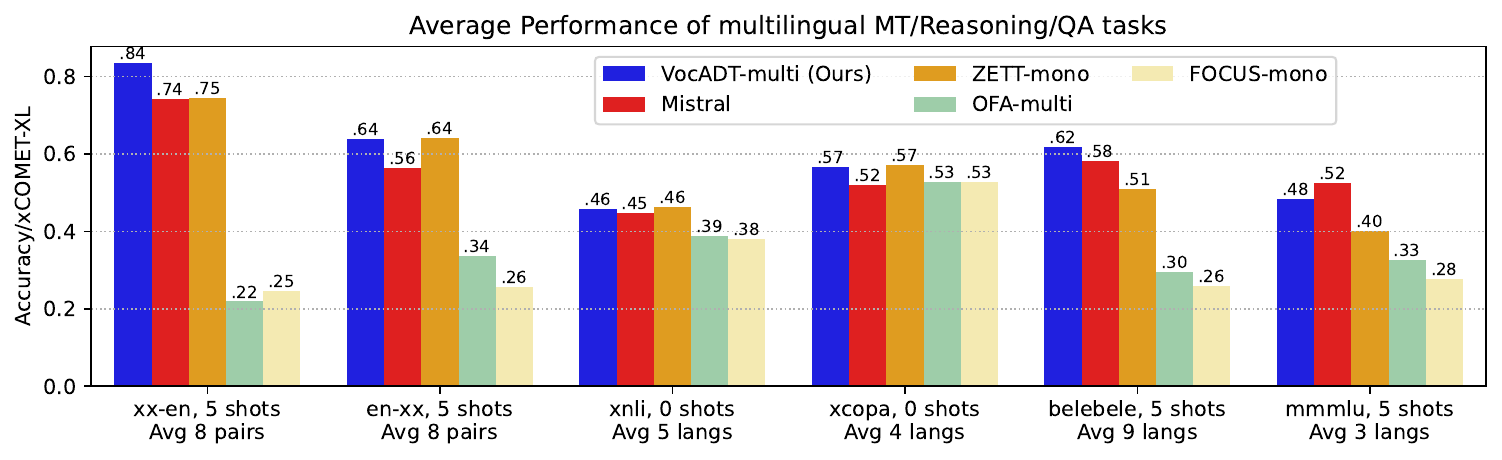} 
\end{center}
\caption{
Average scores of original Mistral and its adaptation with new vocabulary, only replacing embeddings and fixing the body of transformer modules. 
``-multi'' indicates models with a multilingual vocabulary, which includes five languages covering all languages with two separate models, while ``-mono'' refers to monolingual vocabulary models. 
\texttt{xx-en} and \texttt{en-xx} indicate MT tasks. 
See Appendix~\ref{sec:apdx_result_number_langwise} for individual values.
}
\label{fig:avgtasks}
\end{figure}

\subsection{Evaluation}
\label{sec:exp_eval}
We evaluate adaptation methods with multilingual benchmarks of various tasks including MT, natural language inference (NLI), common sense reasoning, and multiple choice question answering (QA).
For MT of English to non-English (\texttt{en-xx}) and non-English to English (\texttt{xx-en}), we use FLORES~\citep{flores101,flores22} as it supports all the languages that we experiment with. 
We use five-shot MT prompting for the model from the adaptation phase, and zero-shot prompting for the model after the ALMA training phase.
We assess the translation quality with xCOMET-XL~\citep{xcomet23}, which produces a score of increasing quality ranging from 0 to 1.
For NLI and reasoning, we use XNLI~\citep{xnli} and XCOPA~\citep{xcopa20} with zero-shot prompting. %
For multiple choice QA, we use Belebele~\citep{belebele23} and Multilingual MMLU~\citep[MMMLU]{mmlu21,okapi23} with five shot prompting. %
All the tasks except for MT are classification tasks, where 
we use the \texttt{lm-evaluation-harness}~\citep{lmeval24} evaluation tool and report accuracy.

\section{Vocabulary Adaptation Results and Analyses}
\label{sec:result_adt}

\subsection{Overall Task Performance}
\label{sec:result_adt_avg}

We first present the controlled comparison on diverse tasks of the original Mistral with new vocabulary variants obtained by our vocabulary adaptation approach (\adt{}) and the ZeTT and OFA baselines.
Figure~\ref{fig:avgtasks} presents the average performance across multiple multilingual MT, NLI, reasoning, and QA tasks. 
Language-wise results are in Appendix \ref{sec:apdx_result_number_langwise}.
Overall, adapting the vocabulary using \adt{} generally leads to better performance compared to the original Mistral model, and either surpasses or performs on par with ZeTT. 
MMMLU is the only task where Mistral still holds the top spot; however, the performance gap between the new and original embeddings is smaller with \adt{} approach than with ZeTT. Remarkably, \adt{}-multi achieves these results with only two models for the eight languages tested, whereas ZeTT requires a separate model for each language.

\begin{figure}[H]
\begin{center}
    \centering
    \begin{subfigure}{\textwidth}
        \centering
        \begin{subfigure}[b]{0.619\textwidth}
            \includegraphics[width=\textwidth]{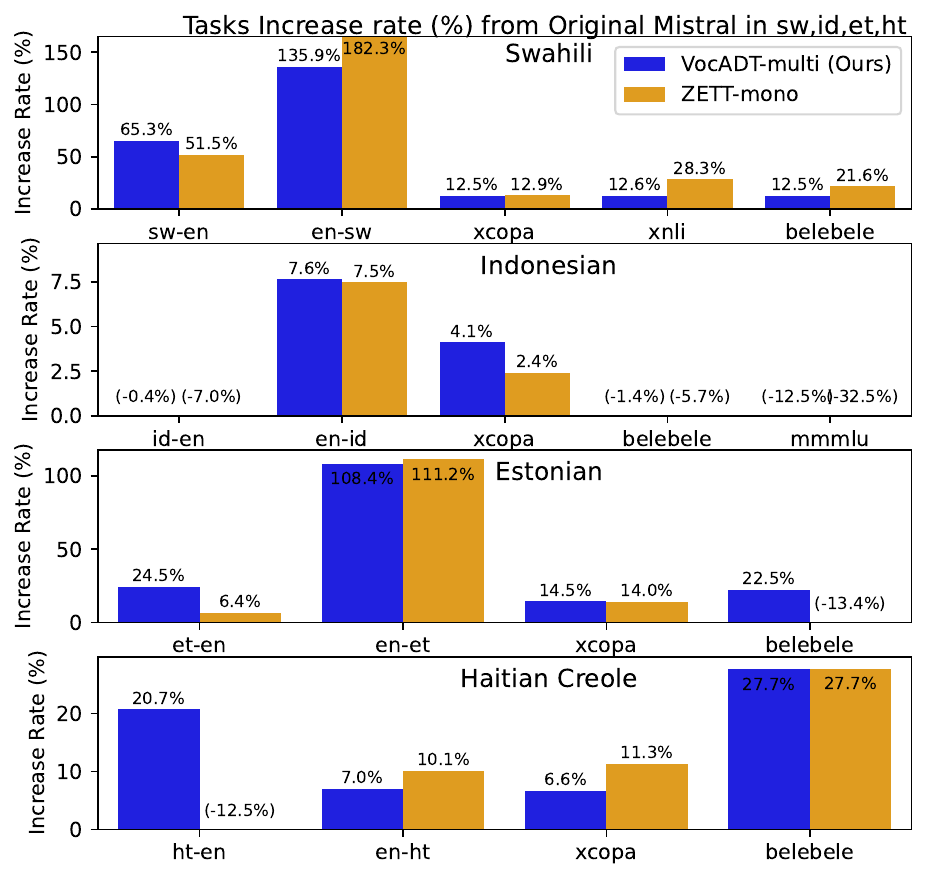}
        \end{subfigure}
        \begin{subfigure}[b]{0.375\textwidth}
            \includegraphics[width=\textwidth]{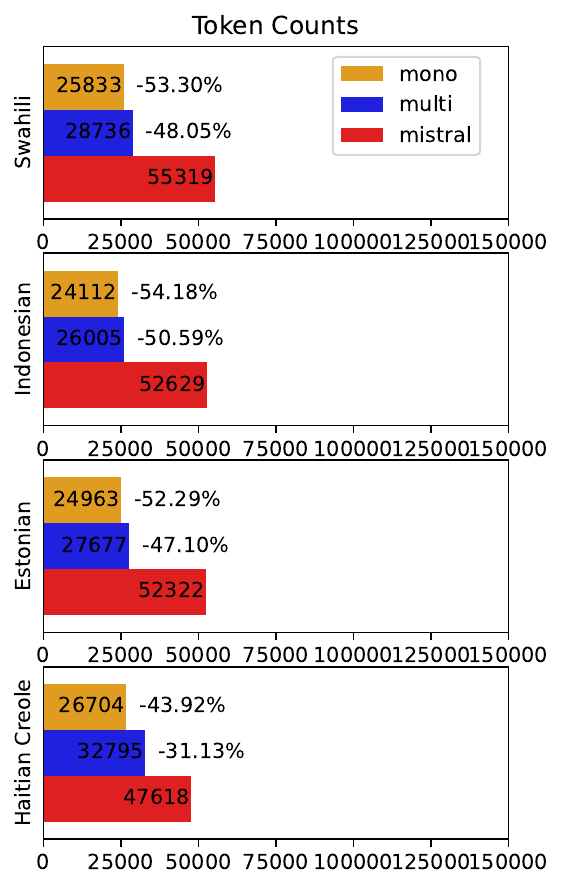}
        \end{subfigure}
        \caption{Increase rate of task performance and number of token count of \textit{Latin} group languages. %
        }
        \label{fig:incre_rate_swidetht}
    \end{subfigure}
    \begin{subfigure}{\textwidth}
        \centering
        \begin{subfigure}[b]{0.619\textwidth}
            \includegraphics[width=\textwidth]{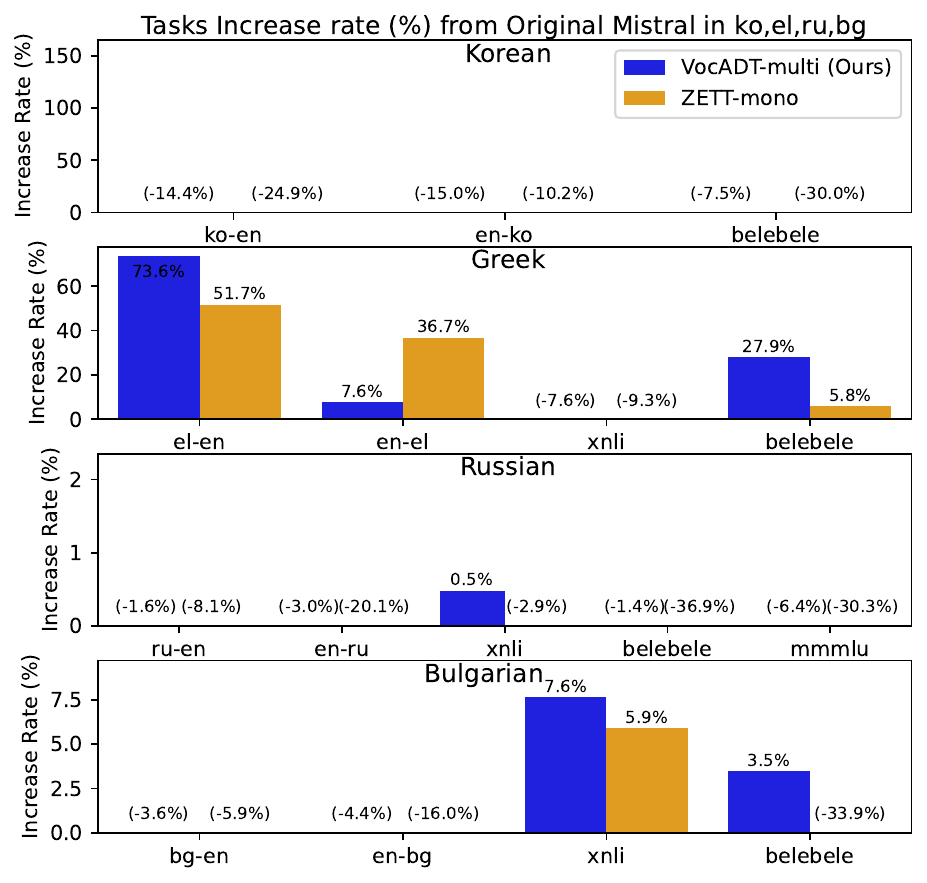}
        \end{subfigure}
        \begin{subfigure}[b]{0.375\textwidth}
            \includegraphics[width=\textwidth]{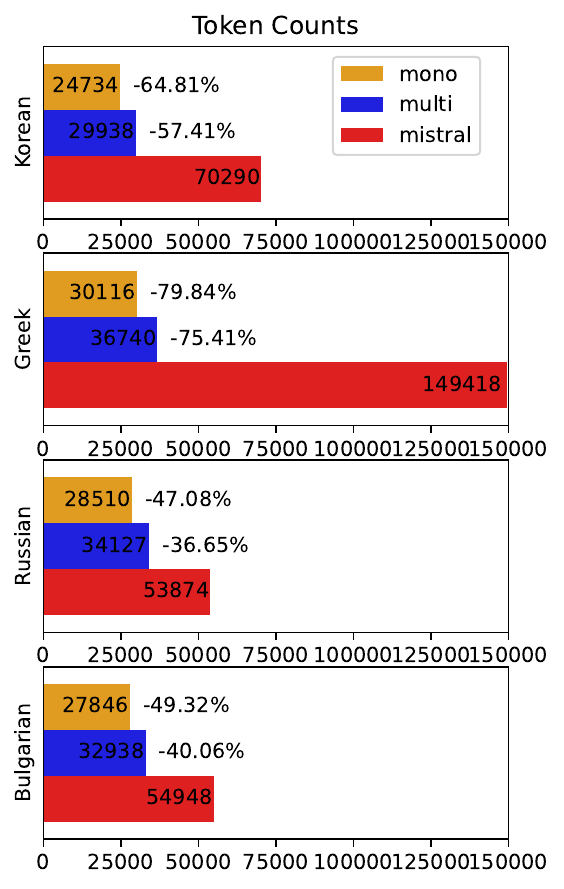}
        \end{subfigure}
        \caption{Increase rate of task performance and number of token count of \textit{Mixed} group languages. %
        }
        \label{fig:incre_rate_koelrubg}
    \end{subfigure}
\end{center}
\caption{
Effect of vocabulary adaption on mitigating over-fragmentation and task performance. The $y$-axis for the increase rate on the left side is limited to the positive range.
Languages with Latin scripts or those experiencing severe fragmentation benefit the most.
\texttt{xx-en} and \texttt{en-xx} are machine translation tasks.
See Appendix~\ref{sec:apdx_result_number_langwise} for individual task performance values.
}
    \label{fig:incre_rate}
\end{figure}

\subsection{Which Languages Benefit the Most from Vocabulary Adaptation?}
\label{sec:result_adt_incre}

The previous section presented a macro view of performance across all languages. This section drills down to look on the impact of vocabulary adaptation in a language-wise manner.
Figure~\ref{fig:incre_rate} shows the increased rate of task performance and tokenization statistics for various languages after applying different vocabulary adaptation methods.
The results are shown for \textit{Latin} group languages (Swahili, Indonesian, Estonian, and Haitian Creole) and \textit{Mixed} group (Korean, Greek, Russian, and Bulgarian). 
We compare \adt{} and ZeTT against the original Mistral model.
The vertical axis of the increase rate is fixed to the positive range to see the benefit trends more easily and all the numbers including the negative range of the increase rate are in Appendix \ref{sec:apdx_result_number_langwise}.
We use the FLORES development set for counting the tokens by various tokenizers where the semantic contents of every language are the same.

\paragraph{Languages with Latin Scripts or Severe Fragmentation Benefit the Most}
In Figures~\ref{fig:incre_rate_swidetht}, we observe that Latin script languages consistently benefit from vocabulary adaptation, regardless of the task, adaptation method employed. %
However, even non-Latin languages show improvements when they suffer from severe over-fragmentation, as seen in the case of Greek in Figures~\ref{fig:incre_rate_koelrubg}. 
Among the eight languages, Greek is the most fragmented by the Mistral tokenizer, and it demonstrates significant improvement after the adaptation to less fragmented vocabulary, particularly in MT tasks, while other non-Latin languages in \textit{Mixed} group show zero, modest, or even negative gains.

In Appendix~\ref{sec:greekkorean_explanation}, we further discuss the pronounced performance declines observed in Korean compared to Russian or Bulgarian within the same Mixed group.
Despite improving fragmentation for Korean, we suspect that the linear combination assumption is insufficient given the lack of representation of Korean characters in the original vocabulary.

\begin{figure}[]
\centering
\begin{subfigure}[b]{0.57\textwidth}
    \includegraphics[width=\textwidth]{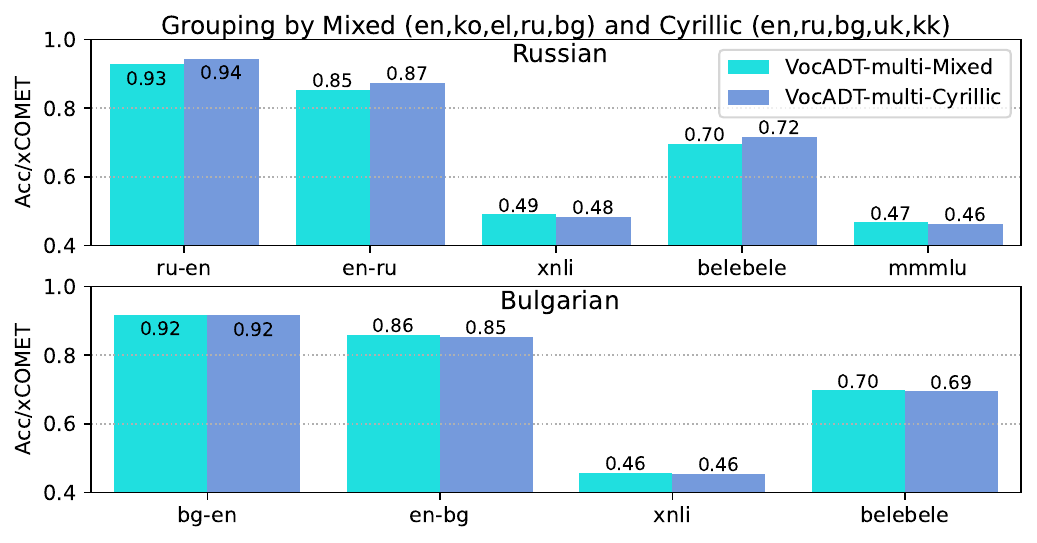}
\end{subfigure}
\begin{subfigure}[b]{0.42\textwidth}
    \includegraphics[width=\textwidth]{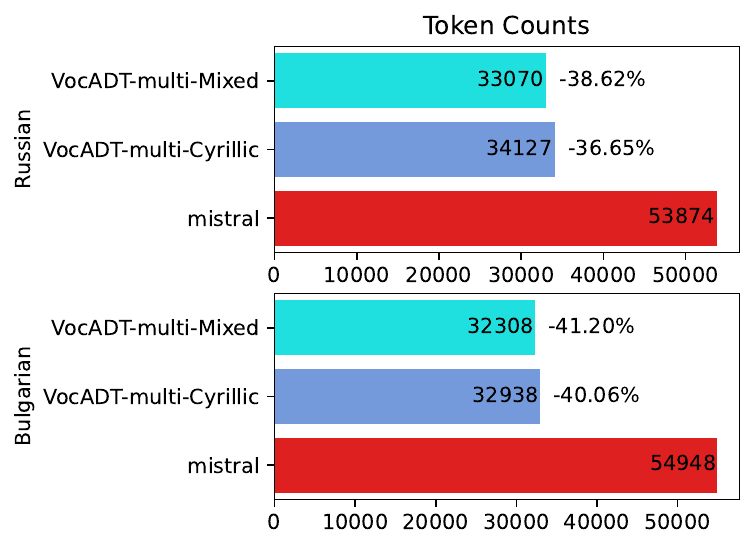}
\end{subfigure}
\caption{Comparison of task performance between two grouping strategies of \textit{Mixed}-script and \textit{Cyrillic}-script on two shared languages. Consistent script within a group provides minor benefits.}
    \label{fig:cyrilic}
\end{figure}

\subsection{Does Script Matter for Language grouping?}
\label{sec:result_adt_cryllic}

Multilingual vocabularies for language groups can strike a balance between the extensive coverage of the original Mistral and the limited scope of language-specific monolingual models.
We investigate strategies for grouping, in particular the effect of script.

Figure~\ref{fig:cyrilic} compares the performance and token count reduction between two non-English grouping strategies for Russian and Bulgarian: \textit{Mixed}-script (\texttt{ko}, \texttt{el}, \texttt{ru}, \texttt{bg}) and \textit{Cyrillic}-script (\texttt{ru}, \texttt{bg}, \texttt{uk}, \texttt{kk}) languages. 
For Russian, the consistent script language group performs slightly better, especially in the MT task.
For Bulgarian, both grouping strategies deliver nearly identical results.
Overall, the results suggest that maintaining a consistent script within a group enhances performance, though outcomes tend to be influenced more by the language type itself than by the grouping strategy.

\subsection{Scalability and Generalizability of \adt{}}
\label{sec:result_all_lma}
We further explore the language scalability of the method with \textit{All} language groups including 11 languages (\textsection\ref{sec:result_all}), and the generalizability of our \adt{} findings to other language models (\textsection\ref{sec:result_lma}). Both scalability and generalizability experiments show that the \textit{All} group follows trends similar to the \textit{Latin}, \textit{Mixed}, and \textit{Cyrillic} setups while the performance trends observed with LLaMA are consistent with those seen in Mistral.

\section{Impact of Vocabulary Adaptation on Downstream Fine-tuning}
\label{sec:result_alma}

Do the effects of vocabulary adaptation hold up after fine-tuning the adapted language model?  After completing the \adt{} process, which keeps non-embedding model weights fixed, we update the full weights of the adapted model to enhance MT performance following ALMA \citep{alma24}. 

As can be seen in Table~\ref{table:mt}, all vocabulary adaptation approaches are effective compared to Mistral except for \texttt{en-sw}, and among those, our approach (\adt{}) achieves the highest average score in both \texttt{en-xx} and \texttt{xx-en} directions.
In the \texttt{xx-en} direction, the performance of \adt{} matches that of ZeTT, despite using a smaller number of individual models of the same size (2 \adt{} vs 8 ZeTT).
Interestingly, language-specific models (ZeTT, FOCUS) tend to excel in Latin languages, whereas multilingual models (Mistral, \adt{}, OFA) generally outperform language-specific models in non-Latin cases.

In sum, with full parameter fine-tuning after the vocabulary adaptation, our \adt{} model offers a competitive edge across both \texttt{xx-en} and \texttt{en-xx} tasks, further validating the effectiveness of our approach.
\adt{} demonstrates that a multilingual model can achieve or surpass language-specific models like ZeTT, offering a more flexible and scalable solution for handling multiple languages.

\input{tables/mt}
\section{Conclusion}
\label{sec:conclusion}
We propose a simple and effective vocabulary adaptation method using a vocabulary adapter.
Our approach consistently outperforms the original Mistral model after the adaptation phase across various tasks and after the following full-weight finetuning on machine translation. 
Furthermore, our method is on par with or more effective than
strong vocabulary adaptation baselines, without relying on external embeddings or language constraints, %
offering a flexible and scalable solution for handling multiple languages.
Our experiments cover a wide range of languages and scripts, revealing that languages with Latin scripts or severe fragmentation benefit the most. %
We also explored different grouping strategies, finding that maintaining consistent scripts within a group offers relatively minor benefits. 
Lastly, with a focus on machine translation, we confirm that vocabulary adaptation remains effective even after full-weight fine-tuning, and \adt{} is the most effective approach.

\section*{Acknowledgements}
We thank Anthony Aue for early discussions, and Marcin Junczys-Dowmunt and the anonymous reviewers for their insightful and helpful feedback.

\bibliography{iclr2025_conference}
\bibliographystyle{iclr2025_conference}

\appendix

\newpage

\section{Is the Auxiliary Loss Helpful?}
\label{sec:result_adt_ovl}

\begin{figure}[h]
\centering
\begin{subfigure}[b]{0.49\textwidth}
    \includegraphics[width=\textwidth]{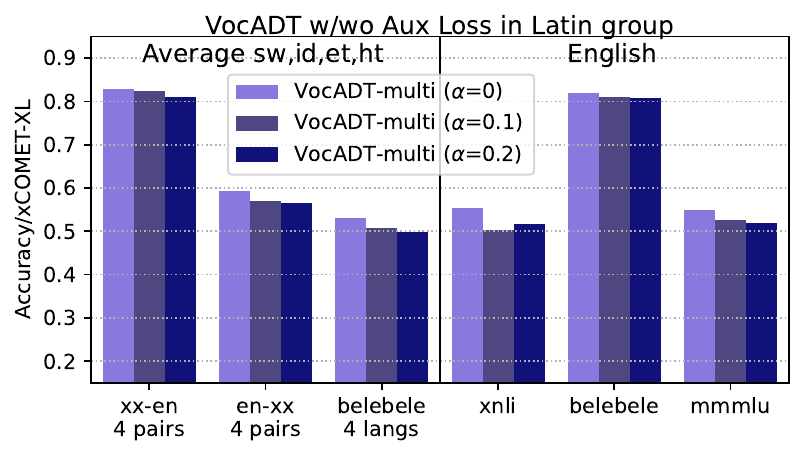}
\end{subfigure}
\begin{subfigure}[b]{0.49\textwidth}
    \includegraphics[width=\textwidth]{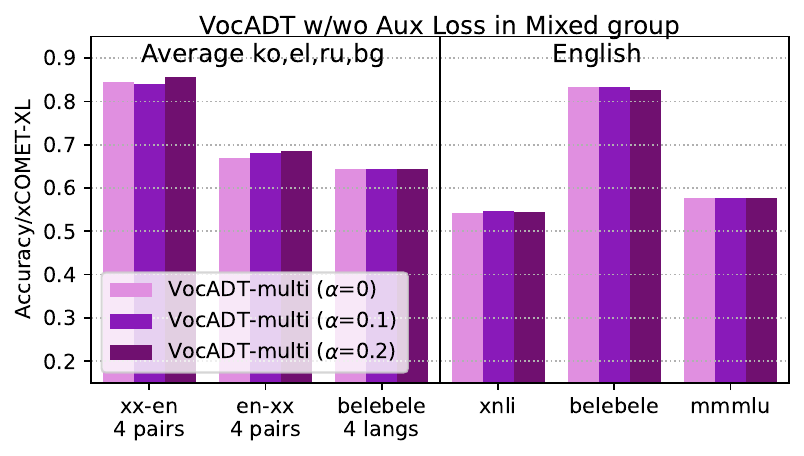}
\end{subfigure}
\caption{
Effects of the auxiliary loss on various settings. 
}
\label{fig:ovl}
\end{figure}

We examine the effects of the auxiliary loss that aims to mitigate the divergence of original status for overlapping words as described in Section~\ref{sec:method_adt_auxloss}.
Figure~\ref{fig:ovl} illustrates the impact on task performance of vocabulary adaptation with and without auxiliary loss on \textit{Latin} and \textit{Mixed} group vocabulary.
We report the average of four non-English languages in each group along with English.

For \textit{Latin} languages (left plot of Figure~\ref{fig:ovl}), 
omitting the auxiliary loss ($\alpha=0$) performs slightly better or comparably to using a non-zero $\alpha$.
For the \textit{Mixed} group plus English vocabulary (right plot of Figure~\ref{fig:ovl}), maintaining the embedding values of overlapping words shows slight effectiveness in both non-English and English. 
We hypothesize that non-Latin languages are less prone to have word collisions with the original vocabulary compared to the \textit{Latin} group, as the Mistral model is largely English (Latin) centric.
As a result, retaining the established embeddings for overlapped words in \textit{Latin} group vocabulary and Mistral vocabulary may disrupt effective learning due to the possible similarity in scripts with English.
On the other hand, keeping the original embeddings during the adaptation for overlapping tokens in \textit{Mixed} may be helpful to maintain the already established embeddings for overlapped tokens while adjusting the embeddings for new non-Latin script tokens.

This auxiliary loss can be thought of as a form of regularization. 
There is a long history of applying regularization during adaptation of MT models to control the adaptation process by limiting the amount that the output distribution or the weights of the fine-tuned model can vary from the original model weights. Prior work has explored using dropout and L2 regularization \citep{miceli-barone-etal-2017-regularization}, cross entropy \citep{khayrallah-etal-2018-regularized}, freezing parts of the network \citep{wuebker-etal-2018-compact, thompson-etal-2018-freezing}, and Elastic Weight Consolidation \citep{doi:10.1073/pnas.1611835114, thompson-etal-2019-overcoming, thompson-etal-2019-hablex}. As our auxiliary loss has mixed effectiveness depending on language characteristics, future work could consider other methods.

\section{Discussions on non-alphabetic scripts and Possible Limitations of Linear Combination Assumption}
\label{sec:greekkorean_explanation}
As shown in Figure~\ref{fig:incre_rate_koelrubg}, the performance decline for Korean in VocADT vocabulary transfer (-14\% to -15\% in MT) is more pronounced than for Russian or Bulgarian (-1\% to -4\%) within the same Mixed group, even though the fragmentation improvement for Korean is greater.
This deviates from the expected performance improvements seen in Greek, where mitigating extreme over-fragmentation (150k tokens) led to gains. Although Korean has significant over-fragmentation (70k tokens), its severity is less than half that of Greek and closer to Russian and Bulgarian (54k tokens).

One possible explanation is the limitation of our assumption that new embeddings can be solely represented as linear combinations of old embeddings. This assumption may not hold well for Korean, which uses the non-alphabetic Hangul script. In Hangul, tokens represent entire syllables or consonant-vowel combinations rather than individual phonemes, making them difficult to decompose into subwords using the original tokenizer.
For instance, the new token ``처럼'' (meaning ``like'' in English) appears frequently in Korean. However, the original Mistral vocabulary lacks a dedicated token for ``럼'', preventing proper decomposition of ``처럼'' without resorting to byte-level tokens. These byte-level fallbacks may not effectively capture the linguistic structure of the character, potentially degrading performance. This issue is less prevalent in alphabetic scripts such as Latin, Cyrillic, or Greek, where words can be easily broken down into individual characters.
This limitation may account for the observed performance discrepancy.

\section{Details of Training, Baseline \& Evaluation}
\label{sec:apdx_training_details}
Here we describe training in detail.
We use four Nvidia A100 GPUs for adapter training and 16 AMD MI200 GPUs for full-weight fine-tuning.
For all monolingual training including adaptation phase and fine-tuning phase, we follow \citep{alma24} for setting the sampling ratio of monolingual data to mitigate language balance in the monolingual data and avoid prioritizing English.
The method fixes the sampling ratio for English with a certain probability e.g. $1 / n$ if there are $n$ languages to mix and allocate the remaining ratio (e.g. $\frac{n-1}{n}$) by employing temperature sampling suggested by \citet{aharoni-etal-2019-massively}.
We mix the monolingual data for \textit{Latin} group $\{$\texttt{en}, \texttt{sw}, \texttt{id}, \texttt{et}, \texttt{ht}$\}$ with $\{$17\%, 16\%, 32\%, 23\%,  12\%$\}$ ratio, for \textit{Mixed} group $\{$\texttt{en}, \texttt{ko}, \texttt{el}, \texttt{ru}, \texttt{bg}$\}$ with $\{$17\%, 17\%, 19\%, 30\%,  17\%$\}$, and for \textit{Cyrillic} $\{$\texttt{en}, \texttt{ru}, \texttt{bg}, \texttt{uk}, \texttt{kk}$\}$ with $\{$17\%, 32\%, 18\%, 20\%,  13\%$\}$.
For \textit{Add} group in \textsection\ref{sec:result_all_lma} and \textsection\ref{sec:result_all}, $\{$\texttt{en}, \texttt{sw}, \texttt{id}, \texttt{et}, \texttt{ht}, \texttt{ko}, \texttt{el}, \texttt{ru}, \texttt{bg}, \texttt{uk}, \texttt{kk}$\}$ with $\{$10\%, 7\%, 10\%, 9\%, 6\%, 9\%, 12\%, 10\%, 9\%, 10\%, 8\%$\}$ ratio.

For parallel training data for the MT task, we use bitext from the NLLB dataset \citep{schwenk-etal-2021-ccmatrix, heffernan-etal-2022-bitext, flores22}\footnote{\url{https://huggingface.co/datasets/allenai/nllb}} This includes web-scraped data, which has the potential to include nosise such as text being automatically identified as the wrong language, mis-aligned or mis-translated segments, and low-quality machine translated segments \citep{khayrallah-koehn-2018-impact,caswell-etal-2020-language, dodge-etal-2021-documenting, 10.1162/tacl_a_00447, thompson-etal-2024-shocking}.
We use LASER3 \citep{laser19} to select higher quality segments for fine-tuning.
LASER has been used extensively to both locate parallel segments on the web \citep{schwenk-etal-2021-wikimatrix, schwenk-etal-2021-ccmatrix} as well as for filtering noisy sentence and document pairs \citep{chaudhary-etal-2019-low, koehn-etal-2020-findings, thompson-koehn-2020-exploiting, sloto-etal-2023-findings}.

In adapter training for \adt{}, we use a (peak) learning rate of 2e-6 with a cosine scheduler, a maximum sequence length of 512 tokens, a warm-up ratio of 0.01, and a weight decay of 0.01.
In full-weight fine-tuning phase, we mostly follow the training setting from ALMA. %
\paragraph{Details of Baseline.}
For ZeTT, we use multilingual hypernetwork for Mistral-7B.\footnote{\url{https://github.com/bminixhofer/zett}} We use the code of OFA\footnote{\url{https://github.com/cisnlp/ofa}} and FOCUS\footnote{\url{https://github.com/konstantinjdobler/focus}} to create new embeddings for Mistral-7B.

\paragraph{Machine Translation Metrics.}
We assess translation quality using xCOMET-XL~\citep{xcomet23}, as recent WMT metric shared tasks \citep{freitag-etal-2023-results, freitag-etal-2024-llms} have found neural metrics like Yisi \citep{lo-2019-yisi, lo-larkin-2020-machine}, Bert-score \citep{bertscorepaper}, Prism \citep{thompson-post-2020-automatic, thompson-post-2020-paraphrase}, Comet \citep{rei-etal-2020-comet},  BLEURT \citep{sellam-etal-2020-bleurt}
correlate much better with human judgements, than surface-form metrics like BLEU \citep{papineni-etal-2002-bleu} or chrF \citep{popovic-2015-chrf, popovic-2017-chrf} which consider only surface form. Trained metrics like the comet and BLEURT, which train on prior human annotations of translation quality, achieve the highest correlation with human judgments. While these correlations are less strong out of domain (relative to the domains used in WMT, e.g. FLORES) the trained metrics still outperform surface level ones \citep{zouhar-etal-2024-fine}.

We also caveat that xCOMET-XL does not consider context when judging translation quality, and context has been shown to be an important aspect of translation quality evaluation (for a thorough overview see \citep{Castilho_Knowles_2024}, especially for LLMs \cite{karpinska-iyyer-2023-large}. While there have been several efforts to incorporate context in MT evaluation (e.g. \cite{vernikos-etal-2022-embarrassingly, deutsch-etal-2023-training, raunak-etal-2024-slide}, there is no consensus in the community as to which method, so we stick to the established xCOMET-XL at the sentence level. Finally, metric differences, especially small ones, may not correspond to statistically significant differences \citep{koehn-2004-statistical, deutsch-etal-2021-statistical, lo-etal-2023-beyond, thompson-etal-2024-improving}.

\section{Computational Cost \adt{}}
\label{sec:appndx_comp}
We report the computational cost of our approach.
We use a batch size of 128 (four A100 GPUs * 8 batch size * 4 gradient accumulation) and a sequence length of 512. The FLOPs per token for \adt{} is 17.7GFLOPs/token, resulting in 1160T ``FLOPs per batch'' (128 * 512 * 17.7G). 
Our training requires 38k update steps (2.49B, roughly 0.5B per langs).
Therefore, the computational cost of a \adt{} model (1160T ``FLOPs per batch'' x 38k step).
We use \texttt{profile()} method of \texttt{accelerator} Python library and our estimation of the FLOPs per token for Mistral-7B is 14.2 GFLOPs/token.

\section{Language-wise Results of Vocabulary Adaptations}
\label{sec:apdx_result_number_langwise}

\input{tables/appndx_adt}

\input{tables/appndx_related_works}

\section{Additional Experiment for Scalability and Generalizability of \adt{}}
\label{sec:appndx_result_all_lma}
\subsection{Combining Languages of \textit{Latin}, \textit{Mixed}, and \textit{Cyrillic} into \textit{All} group}
\label{sec:result_all}
In Section~\ref{sec:result_adt_cryllic}, we observed that while grouping languages for the new vocabulary with a consistent script improves performance, script-based grouping strategies had little overall impact. This suggests that we can enhance the method’s scalability for greater practicality with minimal performance tradeoffs. In this section, we explore a multilingual group with shared vocabulary at larger scales 
to provide better insights into scalability for multilingual setups.

We combine languages from the \textit{Multi}---\textit{Latin}, \textit{Mixed}, and \textit{Cyrillic} ---groups into one unified set into \textit{All}. This set comprises 11 languages---English and 10 non-English languages (Swahili, Indonesian, Estonian, Haitian, Korean, Greek, Russian, Bulgarian, Ukrainian, and Kazakh) as listed in Table~\ref{table:langs}. Following our experimental setup of 0.5B tokens per language, we train on a combined corpus of 5.5B monolingual tokens, covering all 11 languages.\footnote{Available in \url{https://huggingface.co/h-j-han/Mistral-7B-VocADT-50k-All}} We set $\alpha=0$.

Tables~\ref{table:appndx_mt_lma} and~\ref{table:appndx_acc_lma} show that the \textit{All} group follows trends similar to the initial \textit{Latin}, \textit{Mixed}, and \textit{Cyrillic} setups.
Figure~\ref{fig:incre_rate_cmt_all_lma_cr} further illustrates that while the token count for the \textit{All} group is slightly higher than that of the \textit{Multi} group setup, it remains significantly lower than that of the original Mistral model.

\subsection{Generalization to LlaMA}
\label{sec:result_lma}
We primarily conducted our experiments using the Mistral model.
To validate the generalizability of our \adt{} findings to other language models, we also test our approach on an additional candidate LM, LLaMA \citep{touvron2023llama}.

\begin{figure}[h]
\begin{center}
    \centering
    \includegraphics[scale=0.59]{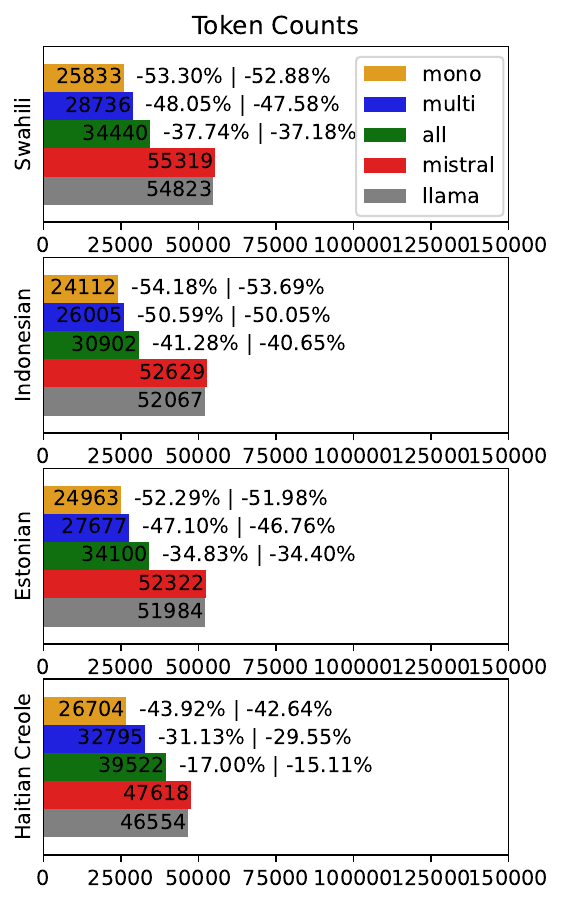}
    \includegraphics[scale=0.59]{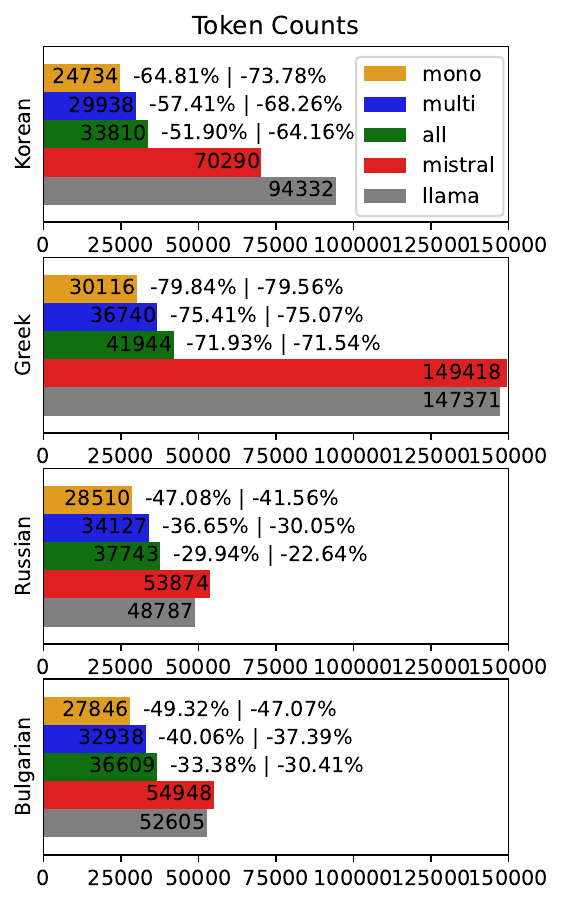}
\end{center}
\caption{
Token count reduction with new vocabulary.
Each bar displays two percentage reduction values: the first (e.g., -53.30\% in Swahili) indicates the reduction relative to the original Mistral model, while the second (e.g., -52.88\%) represents the reduction relative to the original LLaMA model.
We use the FLORES development set for counting the tokens by various tokenizers where the semantic contents of every language are the same.
While \textit{All} group with all 11 languages is slightly higher than that of the \textit{Multi} group with five languages, it remains significantly lower than that of the original models.
The severity of fragmentation in LLaMA is similar to that in Mistral, with Greek being the most severely fragmented language followed by Korean.
}
\label{fig:incre_rate_cmt_all_lma_cr}
\end{figure}

We conducted an additional adaptation experiment using LLaMA2-7B, following the same experimental setup described in the main section.
Figure~\ref{fig:incre_rate_cmt_all_lma_cr} shows that the severity of fragmentation in LLaMA is similar to that in Mistral, with Greek being the most severely fragmented language followed by Korean.
Tables~\ref{table:appndx_mt_lma} and~\ref{table:appndx_acc_lma} confirm that the performance trends observed with LLaMA are consistent with those seen in Mistral.
Overall, Latin group languages benefit largely from vocabulary adaptation, while non-Latin languages in the Mixed group show minus or modest gains, except for Greek, which benefits due to its severe fragmentation.
These findings validate that our method generalizes effectively to another language model.

\input{tables/alllang_llama_result}

\end{document}

%% file: math_commands.tex
\usepackage{amsmath,amsfonts,bm}

\def\eqref#1{equation~\ref{#1}}

\def\1{\bm{1}}

\def\mA{{\bm{A}}}

\def\mE{{\bm{E}}}

\DeclareMathAlphabet{\mathsfit}{\encodingdefault}{\sfdefault}{m}{sl}
\SetMathAlphabet{\mathsfit}{bold}{\encodingdefault}{\sfdefault}{bx}{n}

\newcommand{\R}{\mathbb{R}}



%% file: tables/langs.tex
\begin{table}[t]
\centering
\fontsize{8}{10}\selectfont
\caption{
Covered Languages and its availability in multilingual benchmarks. We mainly categorize non-English languages by scripts---\textit{Latin} group (2-5) and the \textit{Mixed} group (6-7).
We additionally experiment with \textit{Cyrillic} group (8-11).
We follow the resource-level of languages from \citet{joshi-etal-2020-state} and \citet{ustun-etal-2024-aya}
}
\label{table:langs}
\begin{tabular}{llllllllll}
\toprule
idx & Full Name      & Short & Script   & Resource & FLORES        & XNLI      & XCOPA     & Belebele  & MMMLU     \\
\midrule
1 & English        & \texttt{en}    & Latin    & High     & \ding{51} & \ding{51} &           & \ding{51} & \ding{51} \\
\hline 
2 & Swahili        & \texttt{sw}    & Latin    & Low      & \ding{51} & \ding{51} & \ding{51} & \ding{51} &           \\
3 & Indonesian     & \texttt{id}    & Latin    & Mid      & \ding{51} &           & \ding{51} & \ding{51} & \ding{51} \\
4 & Estonian       & \texttt{et}    & Latin    & Mid      & \ding{51} &           & \ding{51} & \ding{51} &           \\
5 & Haitian Creole & \texttt{ht}    & Latin    & Low      & \ding{51} &           & \ding{51} & \ding{51} &           \\
\hline 
6 & Korean         & \texttt{ko}    & Hangul   & High     & \ding{51} &           &           & \ding{51} &           \\
7 & Greek          & \texttt{el}    & Greek    & Mid      & \ding{51} & \ding{51} &           & \ding{51} &           \\
8 & Russian        & \texttt{ru}    & Cyrillic & High     & \ding{51} & \ding{51} &           & \ding{51} & \ding{51} \\
9 & Bulgarian      & \texttt{bg}    & Cyrillic & Mid      & \ding{51} & \ding{51} &           & \ding{51} &           \\
\hline
10 & Ukrainian      &\texttt{uk}    & Cyrillic & Mid      & \ding{51} &           &           & \ding{51} & \ding{51} \\
11 & Kazakh         &\texttt{kk}    & Cyrillic & Mid      & \ding{51} &           &           & \ding{51} &           \\
\bottomrule
\end{tabular}
\end{table}

%% file: tables/mt.tex
\begin{table}[H]
\centering
\fontsize{6.9}{10}\selectfont
\caption{
MT performance after full-weight fine-tuning the new vocabulary-adapted model.
The symbol ``\#''  indicates the number of separate models for this experiment table. %
All vocabulary adaptation approaches after fine-tuning are effective compared to Mistral except for \texttt{en-sw}.
\adt{}-multi shows the best average performance in both directions while matching the score of ZeTT in \texttt{xx-en}. %
}
\label{table:mt}
\begin{tabular}{l|lllll|lllll}
\toprule
FLORES       & \multicolumn{5}{c|}{\texttt{xx-en}}                & \multicolumn{5}{c}{\texttt{en-xx}}                \\
\texttt{Lang} (\textit{group}) $\downarrow$   & \adt{} & Mistral & ZeTT  & OFA   & FOCUS & \adt{} & Mistral & ZeTT  & OFA   & FOCUS \\
\midrule
\texttt{sw} (\textit{Latin})       & 0.893  & 0.891   & \textbf{0.897} & 0.889 & 0.893 & 0.753  & \textbf{0.770}   & 0.762 & 0.763 & 0.762 \\
\texttt{id} (\textit{Latin})       & 0.951  & 0.950   & \textbf{0.953} & 0.951 & 0.945 & 0.874  & 0.874   & \textbf{0.879} & 0.871 & 0.876 \\
\texttt{et} (\textit{Latin})       & 0.939  & 0.925   & \textbf{0.941} & 0.937 & 0.939 & 0.852  & 0.868   & 0.856 & 0.845 & \textbf{0.870} \\
\texttt{ht} (\textit{Latin})       & \textbf{0.706}  & 0.696   & 0.699 & 0.699 & 0.685 & 0.336  & 0.336   & 0.334 & 0.332 & \textbf{0.339} \\
\hline
\texttt{ko} (\textit{Mixed})       & 0.886  & 0.892   & 0.898 & 0.897 & \textbf{0.906} & 0.755  & 0.715   & \textbf{0.772} & 0.742 & 0.754 \\
\texttt{el} (\textit{Mixed})       & \textbf{0.922}  & 0.845   & 0.910 & 0.894 & 0.902 & \textbf{0.876}  & 0.817   & 0.857 & 0.862 & 0.861 \\
\texttt{ru} (\textit{Mixed})       & 0.945  & 0.894   & 0.944 & \textbf{0.946} & 0.945 & \textbf{0.889}  & 0.828   & 0.872 & 0.882 & 0.868 \\
\texttt{bg} (\textit{Mixed})       & \textbf{0.953}  & 0.904   & 0.952 & 0.950 & 0.949 & 0.895  & 0.844   & 0.895 & \textbf{0.899} & 0.895 \\
\hline
Avg (8 pairs)  & \textbf{0.899}  & 0.875   & \textbf{0.899} & 0.895 & 0.895 & \textbf{0.779}  & 0.757   & 0.778 & 0.774 & 0.778 \\
\hline
\texttt{uk} (\textit{Cyrillic})       & \textbf{0.941}  & \textbf{0.941}   & \_     & 0.936 & 0.928 & \textbf{0.878}  & 0.876   & \_     & 0.868 & 0.840 \\
\texttt{kk} (\textit{Cyrillic})       & \textbf{0.881}  & 0.875   & \_     & 0.875 & 0.880 & \textbf{0.807}  & 0.790   & \_     & 0.785 & 0.796 \\
\hline
Avg (10 pairs) & \textbf{0.902}  & 0.881   & \_     & 0.897 & 0.897 & \textbf{0.792}  & 0.772   & \_     & 0.785 & 0.786 \\
\hline
\# of Models $\rightarrow$     &   \textbf{3}    &    \textbf{3}    &   8   & \textbf{3}     &  10   & \textbf{3}    &    \textbf{3}    &   8   & \textbf{3}     &  10 \\
\bottomrule
\end{tabular}
\end{table}

%% file: tables/appndx_adt.tex
\begin{table}[h]
\centering
\fontsize{8}{11}\selectfont
\caption{
\texttt{xx-en} MT results with xCOMET-XL score and the increase rate from the original Mistral after the vocabulary adaptation---only replacing embeddings while fixing the rest.}
\label{table:appndx_mt_xx-en}
\begin{tabular}{l|l|ll|ll|ll|ll}
\toprule
MT (\texttt{xx-en}) & Mistral & \multicolumn{2}{c|}{ \adt{}-multi (Ours)} & \multicolumn{2}{c|}{ZETT-mono} & \multicolumn{2}{c|}{OFA-multi} & \multicolumn{2}{c}{FOCUS-mono} \\
\midrule
\texttt{sw-en}   & 0.485   & 0.801             & 65.32\%  & 0.734     & 51.50\%  & 0.215     & -55.75\% & 0.216      & -55.53\% \\
\texttt{id-en}   & 0.946   & 0.942             & -0.44\%  & 0.880     & -7.01\%  & 0.246     & -73.97\% & 0.186      & -80.36\% \\
\texttt{et-en}   & 0.722   & 0.899             & 24.46\%  & 0.769     & 6.40\%   & 0.196     & -72.93\% & 0.248      & -65.70\% \\
\texttt{ht-en}   & 0.554   & 0.669             & 20.72\%  & 0.484     & -12.55\% & 0.249     & -54.96\% & 0.212      & -61.72\% \\
\texttt{ko-en}   & 0.882   & 0.755             & -14.39\% & 0.662     & -24.87\% & 0.189     & -78.56\% & 0.318      & -63.99\% \\
\texttt{el-en}   & 0.438   & 0.760             & 73.59\%  & 0.664     & 51.69\%  & 0.182     & -58.54\% & 0.250      & -42.87\% \\
\texttt{ru-en}   & 0.959   & 0.927             & -3.33\%  & 0.882     & -8.06\%  & 0.249     & -74.06\% & 0.264      & -72.43\% \\
\texttt{bg-en}   & 0.952   & 0.918             & -3.56\%  & 0.896     & -5.93\%  & 0.228     & -76.00\% & 0.271      & -71.55\% \\
\hline
Avg (8 pairs)  & 0.742   & 0.834             & 12.35\%  & 0.746     & 0.56\%   & 0.219     & -70.47\% & 0.246      & -66.92\% \\
\hline
\texttt{uk-en}   & 0.944   & 0.915             & -3.07\%  & \_        & \_       & 0.201     & -78.70\% & 0.288      & -69.49\% \\
\texttt{kk-en}   & 0.411   & 0.763             & 85.82\%  & \_        & \_       & 0.190     & -53.72\% & 0.308      & -24.98\% \\
\hline
Avg (10 pairs) & 0.729   & 0.835             & 14.49\%  & \_        & \_       & 0.215     & -70.59\% & 0.256      & -64.89\% \\
\midrule
Total \# of Models & \multicolumn{1}{c|}{1}  & \multicolumn{2}{c|}{3} & \multicolumn{2}{c|}{8} & \multicolumn{2}{c|}{3} & \multicolumn{2}{c}{10} \\
\bottomrule
\end{tabular}
\end{table}

\begin{table}[h]
\centering
\fontsize{8}{11}\selectfont
\caption{
\texttt{en-xx} MT results with xCOMET-XL score and the increase rate from the original Mistral after the vocabulary adaptation---only replacing embeddings while fixing the rest.
}
\label{table:appndx_mt_en-xx}
\begin{tabular}{l|l|ll|ll|ll|ll}
\toprule
MT (\texttt{en-xx})     & Mistral & \multicolumn{2}{c|}{ \adt{}-multi (Ours)} & \multicolumn{2}{c|}{ZETT-mono} & \multicolumn{2}{c|}{OFA-multi} & \multicolumn{2}{c}{FOCUS-mono} \\
\midrule
\texttt{en-sw}   & 0.238   & 0.562            & 135.88\%           & 0.673        & 182.29\%       & 0.342        & 43.54\%        & 0.209        & -12.23\%        \\
\texttt{en-id}   & 0.778   & 0.837            & 7.65\%             & 0.836        & 7.47\%         & 0.436        & -43.99\%       & 0.224        & -71.26\%        \\
\texttt{en-et}   & 0.309   & 0.643            & 108.37\%           & 0.652        & 111.23\%       & 0.405        & 31.12\%        & 0.247        & -19.86\%        \\
\texttt{en-ht}   & 0.308   & 0.329            & 7.03\%             & 0.339        & 10.11\%        & 0.227        & -26.38\%       & 0.235        & -23.61\%        \\
\texttt{en-ko}   & 0.703   & 0.598            & -14.99\%           & 0.631        & -10.24\%       & 0.309        & -56.06\%       & 0.258        & -63.33\%        \\
\texttt{en-el}   & 0.384   & 0.413            & 7.56\%             & 0.524        & 36.71\%        & 0.232        & -39.58\%       & 0.215        & -43.90\%        \\
\texttt{en-ru}   & 0.900   & 0.854            & -5.17\%            & 0.719        & -20.10\%       & 0.388        & -56.87\%       & 0.371        & -58.80\%        \\
\texttt{en-bg}   & 0.899   & 0.859            & -4.43\%            & 0.755        & -16.02\%       & 0.332        & -63.03\%       & 0.289        & -67.80\%        \\
\hline
Avg (8 pairs)  & 0.565   & 0.637            & 12.77\%            & 0.641        & 13.53\%        & 0.334        & -40.89\%       & 0.256        & -54.66\%        \\
\hline
\texttt{en-uk}   & 0.865   & 0.851            & -1.59\%            & \_           & \_             & 0.318        & -63.23\%       & 0.310        & -64.14\%        \\
\texttt{en-kk}   & 0.222   & 0.522            & 135.11\%           & \_           & \_             & 0.294        & 32.25\%        & 0.223        & 0.65\%          \\
\hline
Avg (10 pairs) & 0.560   & 0.647            & 15.40\%            & \_           & \_             & 0.328        & -41.44\%       & 0.258        & -53.93\%       \\
\midrule
Total \# of Models & \multicolumn{1}{c|}{1}  & \multicolumn{2}{c|}{3} & \multicolumn{2}{c|}{8} & \multicolumn{2}{c|}{3} & \multicolumn{2}{c}{10} \\
\bottomrule
\end{tabular}
\end{table}

\begin{samepage}
\begin{table}[H]
\centering
\fontsize{8}{10}\selectfont
\caption{
XNLI results with Accuracy score and the increase rate from the original Mistral after the vocabulary adaptation---only replacing embeddings while fixing the rest.
}
\label{table:appndx_xnli}
\begin{tabular}{l|l|ll|ll|ll|ll}
\toprule
XNLI   & Mistral & \multicolumn{2}{c|}{ \adt{}-multi (Ours)} & \multicolumn{2}{c|}{ZETT-mono} & \multicolumn{2}{c|}{OFA-multi} & \multicolumn{2}{c}{FOCUS-mono} \\
\midrule
\texttt{en}     & 0.550   & 0.553            & 0.47\%             & 0.554        & 0.73\%         & 0.547        & -0.47\%        & 0.537        & -2.30\%         \\
\texttt{sw}     & 0.353   & 0.398            & 12.63\%            & 0.453        & 28.33\%        & 0.345        & -2.16\%        & 0.325        & -7.96\%         \\
\texttt{el}     & 0.419   & 0.387            & -7.60\%            & 0.380        & -9.31\%        & 0.330        & -21.21\%       & 0.337        & -19.58\%        \\
\texttt{ru}     & 0.488   & 0.490            & 0.48\%             & 0.474        & -2.87\%        & 0.347        & -28.98\%       & 0.331        & -32.27\%        \\
\texttt{bg}     & 0.425   & 0.457            & 7.63\%             & 0.450        & 5.88\%         & 0.344        & -19.02\%       & 0.371        & -12.59\%        \\
\hline
Avg (5 langs) & 0.447   & 0.457            & 2.24\%             & 0.462        & 3.40\%         & 0.383        & -14.38\%       & 0.380        & -14.93\%       \\
\midrule
Total \# of Models & \multicolumn{1}{c|}{1}  & \multicolumn{2}{c|}{2} & \multicolumn{2}{c|}{5} & \multicolumn{2}{c|}{2} & \multicolumn{2}{c}{5} \\
\bottomrule
\end{tabular}
\end{table}
\begin{table}[H]
\centering
\fontsize{8}{10}\selectfont
\caption{
XCOPA results with Accuracy score and the increase rate from the original Mistral after the vocabulary adaptation---only replacing embeddings while fixing the rest.
}
\label{table:appndx_xcopa}
\begin{tabular}{l|l|ll|ll|ll|ll}
\toprule
XCOPA         & Mistral & \multicolumn{2}{c|}{ \adt{}-multi (Ours)} & \multicolumn{2}{c|}{ZETT-mono} & \multicolumn{2}{c|}{OFA-multi} & \multicolumn{2}{c}{FOCUS-mono} \\
\midrule
\texttt{sw}            & 0.510   & 0.574            & 12.55\%            & 0.576        & 12.94\%        & 0.564        & 10.59\%        & 0.544        & 6.67\%          \\
\texttt{id}            & 0.584   & 0.608            & 4.11\%             & 0.598        & 2.40\%         & 0.508        & -13.01\%       & 0.512        & -12.33\%        \\
\texttt{et}            & 0.470   & 0.538            & 14.47\%            & 0.536        & 14.04\%        & 0.516        & 9.79\%         & 0.520        & 10.64\%         \\
\texttt{ht}            & 0.514   & 0.548            & 6.61\%             & 0.572        & 11.28\%        & 0.526        & 2.33\%         & 0.534        & 3.89\%          \\
\hline
Avg (4 langs) & 0.520   & 0.567            & 9.14\%             & 0.571        & 9.82\%         & 0.529        & 1.73\%         & 0.528        & 1.54\%         \\
\midrule
Total \# of Models & \multicolumn{1}{c|}{1}  & \multicolumn{2}{c|}{1} & \multicolumn{2}{c|}{4} & \multicolumn{2}{c|}{1} & \multicolumn{2}{c}{5} \\
\bottomrule
\end{tabular}
\end{table}

\begin{table}[H]
\centering
\fontsize{8}{10}\selectfont
\caption{
Belebele results with Accuracy score and the increase rate from the original Mistral after the vocabulary adaptation---only replacing embeddings while fixing the rest.
}
\label{table:appndx_belebele}
\begin{tabular}{l|l|ll|ll|ll|ll}
\toprule
Belebele & Mistral & \multicolumn{2}{c|}{ \adt{}-multi (Ours)} & \multicolumn{2}{c|}{ZETT-mono} & \multicolumn{2}{c|}{OFA-multi} & \multicolumn{2}{c}{FOCUS-mono} \\
\midrule
\texttt{en}       & 0.843   & 0.833            & -1.18\%            & 0.780        & -7.51\%        & 0.546        & -35.31\%       & 0.367        & -56.52\%        \\
\texttt{sw}       & 0.391   & 0.440            & 12.50\%            & 0.476        & 21.61\%        & 0.248        & -36.65\%       & 0.252        & -35.51\%        \\
\texttt{id}       & 0.647   & 0.638            & -1.38\%            & 0.610        & -5.67\%        & 0.289        & -55.33\%       & 0.230        & -64.43\%        \\
\texttt{et}       & 0.439   & 0.538            & 22.53\%            & 0.380        & -13.42\%       & 0.250        & -43.04\%       & 0.213        & -51.39\%        \\
\texttt{ht}       & 0.397   & 0.507            & 27.72\%            & 0.507        & 27.73\%        & 0.248        & -37.54\%       & 0.240        & -39.50\%        \\
\texttt{ko}       & 0.666   & 0.616            & -7.52\%            & 0.466        & -30.05\%       & 0.278        & -58.27\%       & 0.274        & -58.77\%        \\
\texttt{el}       & 0.442   & 0.566            & 27.90\%            & 0.468        & 5.79\%         & 0.287        & -35.17\%       & 0.284        & -35.68\%        \\
\texttt{ru}       & 0.727   & 0.696            & -4.29\%            & 0.459        & -36.85\%       & 0.248        & -65.90\%       & 0.239        & -67.13\%        \\
\texttt{bg}       & 0.674   & 0.698            & 3.47\%             & 0.446        & -33.93\%       & 0.276        & -59.14\%       & 0.233        & -65.40\%        \\
\hline
Avg (9 langs)   & 0.581   & 0.614            & 5.83\%             & 0.510        & -12.16\%       & 0.296        & -48.95\%       & 0.259        & -55.35\%        \\
\hline
\texttt{uk}       & 0.728   & 0.693            & -4.76\%            & \_           & \_             & 0.254        & -65.05\%       & 0.231        & -68.25\%        \\
\texttt{kk}       & 0.364   & 0.442            & 21.36\%            & \_           & \_             & 0.256        & -29.87\%       & 0.220        & -39.63\%        \\
\hline
Avg (11 langs)  & 0.574   & 0.606            & 5.50\%             & \_           & \_             & 0.289        & -49.70\%       & 0.253        & -55.93\%       \\
\midrule
Total \# of Models & \multicolumn{1}{c|}{1}  & \multicolumn{2}{c|}{3} & \multicolumn{2}{c|}{9} & \multicolumn{2}{c|}{3} & \multicolumn{2}{c}{11} \\
\bottomrule
\end{tabular}
\end{table}

\begin{table}[H]
\centering
\fontsize{8}{10}\selectfont
\caption{
Multilingual MMLU results with Accuracy score and the increase rate from the original Mistral after the vocabulary adaptation---only replacing embeddings while fixing the rest.
}
\label{table:appndx_mmmlu}
\begin{tabular}{l|l|ll|ll|ll|ll}
\toprule
MMMLU         & Mistral & \multicolumn{2}{c}{ \adt{}-multi (Ours)} & \multicolumn{2}{c}{ZETT-mono} & \multicolumn{2}{c}{OFA-multi} & \multicolumn{2}{c}{FOCUS-mono} \\
\midrule
\texttt{en}            & 0.607   & 0.577             & -4.88\%  & 0.537     & -11.50\% & 0.464     & -23.48\% & 0.288      & -52.61\% \\
\texttt{id}            & 0.468   & 0.410             & -12.49\% & 0.316     & -32.53\% & 0.256     & -45.34\% & 0.269      & -42.60\% \\
\texttt{ru}            & 0.500   & 0.468             & -6.39\%  & 0.348     & -30.34\% & 0.259     & -48.25\% & 0.272      & -45.55\% \\
\hline
Avg (3 langs) & 0.525   & 0.485             & -7.62\%  & 0.400     & -23.73\% & 0.326     
& -37.84\% & 0.276      & -47.39\% \\
\hline
\texttt{uk}            & 0.489   & 0.462             & -5.57\%  & \_        & \_       & 0.269     & -45.06\% & 0.253      & -48.19\% \\
\hline
Avg (4 langs) & 0.516   & 0.479             & -7.14\%  & \_        & \_       & 0.312     & -39.55\% & 0.270      & -47.58\% \\ 
\midrule
Total \# of Models & \multicolumn{1}{c|}{1}  & \multicolumn{2}{c|}{3} & \multicolumn{2}{c|}{3} & \multicolumn{2}{c|}{3} & \multicolumn{2}{c}{4} \\
\bottomrule
\end{tabular}
\end{table}
\end{samepage}

%% file: tables/appndx_related_works.tex

\begin{table}[h]
\centering
\fontsize{7.9}{10}\selectfont
\caption{
Tables of various vocabulary adaptation methods. The works in \textbf{bold} linearly combine original embeddings to generate new embeddings.
}
\label{table:appndx_related_works}
\begin{tabular}{l|lllll}
\toprule
\begin{tabular}[c]{@{}l@{}}Vocabulary\\ Adaptation\end{tabular}               & Grouping                                                                                           & \begin{tabular}[c]{@{}l@{}}\#\\ Langs\end{tabular} & \begin{tabular}[c]{@{}l@{}}External\\ Resources\end{tabular}                 & \begin{tabular}[c]{@{}l@{}}Base\\ Model\end{tabular}                              & \begin{tabular}[c]{@{}l@{}}Generative\\ Task\end{tabular}   \\
\midrule
\textbf{\adt}~(Ours)                                                                   & \begin{tabular}[c]{@{}l@{}}multilingual\\ (\textit{Latin}, \textit{Mixed}, \\ \textit{Cyrillic} group)\end{tabular}           & 11                                                 & x                                                                            & Mistral                                                                           & MT                                                          \\
\hline
\begin{tabular}[c]{@{}l@{}}ZeTT\\ \citep{zett24}\end{tabular}                 & lang-specific                                                                                      & 26                                                 & x                                                                            & \begin{tabular}[c]{@{}l@{}}Mistral, \\ XLM-R\end{tabular}                         & x                                                           \\
\hline
\begin{tabular}[c]{@{}l@{}}\textbf{RAMEN}\\ \citep{ramen20}\end{tabular}               & lang-specific                                                                                      & 6                                                  & \begin{tabular}[c]{@{}l@{}}FastAlign,\\ fastText\end{tabular}                & \begin{tabular}[c]{@{}l@{}}BERT,\\ RoBERTa\end{tabular}                           & x                                                           \\
\hline
\begin{tabular}[c]{@{}l@{}}\textbf{FVT}\\ \citep{fvt22}\end{tabular}                   & \begin{tabular}[c]{@{}l@{}}English,\\ domain-specific\end{tabular}                                 & 1 (\texttt{en})                                             & x                                                                            & BERT                                                                              & x                                                           \\
\hline
\begin{tabular}[c]{@{}l@{}}\textbf{VIPI}\\ \citep{vipi22}\end{tabular}                 & \begin{tabular}[c]{@{}l@{}}English,\\ domain-specific\end{tabular}                                 & 1 (\texttt{en})                                             & x                                                                            & BERT                                                                              & x                                                           \\
\hline
\begin{tabular}[c]{@{}l@{}}\textbf{OFA}\\ \citep{ofa24}\end{tabular}                   & \begin{tabular}[c]{@{}l@{}}multilingual\\ (all in 401k)\end{tabular}                               & min 369                                            & ColexNet+                                                                    & \begin{tabular}[c]{@{}l@{}}XLM-R, \\ RoBERTa\end{tabular}                         & x                                                                                 \\
\hline
\begin{tabular}[c]{@{}l@{}}\textbf{FOCUS}\\ \citep{focus23}\end{tabular}               & lang-specific                                                                                      & 10                                                 & fastText                                                                     & XLM-R                                                                             & x                                                           \\
\hline
\begin{tabular}[c]{@{}l@{}}MAD-X\\ \citep{madx20}\end{tabular}                & lang-specific                                                                                      & 16                                                 & x                                                                            & XLM-R                                                                             & x                                                           \\
\hline
\begin{tabular}[c]{@{}l@{}}\textbf{WECHSEL}\\ \citep{wechsel22}\end{tabular}           & lang-specific                                                                                      & 8                                                  & \begin{tabular}[c]{@{}l@{}}fastText,\\ bilingual\\ dictionaries\end{tabular} & \begin{tabular}[c]{@{}l@{}}RoBERTa, \\ GPT-2\end{tabular}                         & x                                                           \\
\hline
\begin{tabular}[c]{@{}l@{}}\textbf{CW2V}\\ \citep{cw2v24}\end{tabular}                 & \begin{tabular}[c]{@{}l@{}}multilingual\\ (all 4)\end{tabular}                                     & 4                                                  & \begin{tabular}[c]{@{}l@{}}bilingual\\ dictionaries\end{tabular}             & \begin{tabular}[c]{@{}l@{}}LLaMA2, \\ RoBERTa\end{tabular}                        & \begin{tabular}[c]{@{}l@{}}MT,\\ summarization\end{tabular} \\
\hline
\begin{tabular}[c]{@{}l@{}}\textbf{CLP}\\ \citep{clp23}\end{tabular}                   & lang-specific                                                                                      & 1 (\texttt{de})                                             & \begin{tabular}[c]{@{}l@{}}GPT2-base w\\ WECHSEL\end{tabular}                & \begin{tabular}[c]{@{}l@{}}GPT2, \\ BLOOM\end{tabular}                            & x                                                           \\
\hline
\begin{tabular}[c]{@{}l@{}}\textbf{CLP+}\\ \citep{yamaguchi2024empirical}\end{tabular} & lang-specific                                                                                      & 4                                                  & \begin{tabular}[c]{@{}l@{}}GPT2-base w\\ WECHSEL\end{tabular}                & \begin{tabular}[c]{@{}l@{}}BLOOM-1/7B,\\ TigerBot- 7B, \\ Mistral-7B\end{tabular} & summarization                                               \\
\hline
\citet{downey-etal-2023-embedding}                                            & \begin{tabular}[c]{@{}l@{}}lang-specific \\ \& multilingual \\ (Mixed, Uralic family)\end{tabular} & 10                                                 & x                                                                            & XLM-R                                                                             & x                                                          \\
\bottomrule

\end{tabular}
\end{table}

%% file: tables/alllang_llama_result.tex
\begin{table}[h]
\centering
\fontsize{8}{11}\selectfont
\caption{
\texttt{xx-en} and \texttt{en-xx} MT results with xCOMET-XL score and the increase rate from the original Mistral after the vocabulary adaptation---only replacing embeddings while fixing the rest.
The tables compare the \textit{All} 11-language group versus the \textit{Multi} groups---\textit{Latin}, \textit{Mixed}, and \textit{Cyrillic} (each comprising 5 languages). We also compare the experiments using Mistral versus LLaMA as the base model.
}
\label{table:appndx_mt_lma}
\begin{tabular}{l|l|ll|ll|l|ll}
\toprule
\multirow{2}{*}{MT \texttt{xx-en}} & \multicolumn{5}{c|}{Mistral}                                                                 & \multicolumn{3}{c}{\textbf{Llama}}                       \\ 
                    & \multicolumn{1}{c}{Orig}  & \multicolumn{2}{c}{\adt{}-multi} & \multicolumn{2}{c|}{\adt{}-\textbf{all}} & \multicolumn{1}{c}{Orig}   & \multicolumn{2}{c}{\adt{}-multi} \\
\midrule
Total \# of Models & \multicolumn{1}{c|}{1}       & \multicolumn{2}{c|}{3}                                      & \multicolumn{2}{c|}{\textbf{1}}                                    & \multicolumn{1}{c|}{1}     & \multicolumn{2}{c}{3}            \\
\midrule
\texttt{sw-en}              & 0.485   & 0.801                      & 65.32\%                       & 0.775                      & 59.89\%                     & 0.359 & 0.698                     & 94.43\%                     \\
\texttt{id-en}              & 0.946   & 0.942                      & -0.44\%                       & 0.919                      & -2.89\%                     & 0.954 & 0.933                     & -2.20\%                     \\
\texttt{et-en}              & 0.722   & 0.899                      & 24.46\%                       & 0.851                      & 17.79\%                     & 0.496 & 0.858                     & 72.98\%                     \\
\texttt{ht-en}              & 0.554   & 0.669                      & 20.72\%                       & 0.63                       & 13.74\%                     & 0.392 & 0.645                     & 64.54\%                     \\
\texttt{ko-en}              & 0.882   & 0.755                      & -14.39\%                      & 0.834                      & -5.41\%                     & 0.872 & 0.776                     & -11.01\%                    \\
\texttt{el-en}              & 0.438   & 0.760                      & 73.59\%                       & 0.856                      & 95.44\%                     & 0.439 & 0.777                     & 76.99\%                     \\
\texttt{ru-en}              & 0.959   & 0.927                      & -3.33\%                       & 0.929                      & -3.14\%                     & 0.951 & 0.93                      & -2.21\%                     \\
\texttt{bg-en}              & 0.952   & 0.918                      & -3.56\%                       & 0.92                       & -3.38\%                     & 0.941 & 0.916                     & -2.66\%                     \\
\hline
Avg (8 pairs)             & 0.742   & 0.834                      & 12.35\%                       & 0.839                      & 13.03\%                     & 0.675 & 0.817                     & 21.04\%                     \\
\hline
\texttt{uk-en}              & 0.944   & 0.915                      & -3.07\%                       & 0.909                      & -3.74\%                     & 0.947 & 0.897                     & -5.28\%                     \\
\texttt{kk-en}              & 0.411   & 0.763                      & 85.82\%                       & 0.751                      & 82.92\%                     & 0.286 & 0.611                     & 113.64\%                    \\
\hline
Avg (10 pairs)            & 0.729   & 0.835                      & 14.49\%                       & 0.837                      & 14.76\%                     & 0.664 & 0.804                     & 21.08\%                     \\
\toprule
\multirow{2}{*}{MT \texttt{en-xx}} & \multicolumn{5}{c|}{Mistral}                                                                 & \multicolumn{3}{c}{\textbf{Llama}}                       \\ 
                    & \multicolumn{1}{c}{Orig}  & \multicolumn{2}{c}{\adt{}-multi} & \multicolumn{2}{c|}{\adt{}-\textbf{all}} & \multicolumn{1}{c}{Orig}   & \multicolumn{2}{c}{\adt{}-multi} \\
\midrule
\texttt{en-sw}              & 0.238   & 0.562                      & 135.88\%                      & 0.466                     & 95.48\%                      & 0.291 & 0.367                     & 26.12\%                     \\
\texttt{en-id}              & 0.778   & 0.837                      & 7.65\%                        & 0.763                     & -1.89\%                      & 0.868 & 0.872                     & 0.46\%                      \\
\texttt{en-et}              & 0.309   & 0.643                      & 108.37\%                      & 0.587                     & 90.12\%                      & 0.279 & 0.581                     & 108.24\%                    \\
\texttt{en-ht}              & 0.308   & 0.329                      & 7.03\%                        & 0.312                     & 1.40\%                       & 0.286 & 0.315                     & 10.14\%                     \\
\texttt{en-ko}              & 0.703   & 0.598                      & -14.99\%                      & 0.631                     & -10.23\%                     & 0.669 & 0.566                     & -15.40\%                    \\
\texttt{en-el}              & 0.384   & 0.413                      & 7.56\%                        & 0.635                     & 65.56\%                      & 0.297 & 0.511                     & 72.05\%                     \\
\texttt{en-ru}              & 0.900   & 0.854                      & -5.17\%                       & 0.855                     & -5.02\%                      & 0.877 & 0.824                     & -6.04\%                     \\
\texttt{en-bg}              & 0.899   & 0.859                      & -4.43\%                       & 0.854                     & -4.96\%                      & 0.826 & 0.825                     & -0.12\%                     \\
\hline
Avg (8 pairs)             & 0.565   & 0.637                      & 12.77\%                       & 0.638                     & 12.98\%                      & 0.549 & 0.608                     & 10.75\%                     \\
\hline
\texttt{en-uk}              & 0.865   & 0.851                      & -1.59\%                       & 0.83                      & -4.05\%                      & 0.83  & 0.814                     & -1.93\%                     \\
\texttt{en-kk}              & 0.222   & 0.522                      & 135.11\%                      & 0.555                     & 150.05\%                     & 0.188 & 0.354                     & 88.30\%                     \\
\hline
Avg (10 pairs)            & 0.560   & 0.647                      & 15.40\%                       & 0.649                     & 15.79\%                      & 0.541 & 0.603                     & 11.46\%                     \\
\bottomrule
\end{tabular}
\end{table}
\begin{table}[h]
\centering
\fontsize{8}{10}\selectfont
\caption{
XNLI, XCOPA, Belebele, and MMMLU results with Accuracy score and the increase rate from the original Mistral after the vocabulary adaptation---only replacing embeddings while fixing the rest. The tables compare the \textit{All} 11-language group versus the \textit{Multi} groups---\textit{Latin}, \textit{Mixed}, and \textit{Cyrillic} (each comprising 5 languages). We also compare the experiments using Mistral versus LLaMA as the base model.
}
\label{table:appndx_acc_lma}
\begin{tabular}{l|l|ll|ll|l|ll}
\toprule
\multirow{2}{*}{XNLI} & \multicolumn{5}{c|}{Mistral}                                                                 & \multicolumn{3}{c}{\textbf{Llama}}                       \\ 
                    & \multicolumn{1}{c}{Orig}  & \multicolumn{2}{c}{\adt{}-multi} & \multicolumn{2}{c|}{\adt{}-\textbf{all}} & \multicolumn{1}{c}{Orig}   & \multicolumn{2}{c}{\adt{}-multi} \\
\midrule
Total \# of Models & \multicolumn{1}{c|}{1}       & \multicolumn{2}{c|}{3}                                      & \multicolumn{2}{c|}{\textbf{1}}                                    & \multicolumn{1}{c|}{1}     & \multicolumn{2}{c}{3}                       \\
\midrule
\texttt{en}                 & 0.550   & 0.553                       & 0.47\%                       & 0.53                       & -3.64\%                     & 0.554 & 0.568                     & 2.53\%                      \\
\texttt{sw}                 & 0.353   & 0.398                       & 12.63\%                      & 0.397                      & 12.46\%                     & 0.348 & 0.378                     & 8.62\%                      \\
\texttt{el}                 & 0.419   & 0.387                       & -7.60\%                      & 0.396                      & -5.49\%                     & 0.370 & 0.382                     & 3.24\%                      \\
\texttt{ru}                 & 0.488   & 0.490                       & 0.48\%                       & 0.494                      & 1.23\%                      & 0.425 & 0.47                      & 10.59\%                     \\
\texttt{bg}                 & 0.425   & 0.457                       & 7.63\%                       & 0.435                      & 2.35\%                      & 0.424 & 0.388                     & -8.49\%                     \\
\hline
Avg (5 langs)             & 0.447   & 0.457                       & 2.24\%                       & 0.451                      & 0.89\%                      & 0.424 & 0.437                     & 3.00\%                      \\
\toprule
\multirow{2}{*}{XCOPA} & \multicolumn{5}{c|}{Mistral}                                                                 & \multicolumn{3}{c}{\textbf{Llama}}                       \\ 
                    & \multicolumn{1}{c}{Orig}  & \multicolumn{2}{c}{\adt{}-multi} & \multicolumn{2}{c|}{\adt{}-\textbf{all}} & \multicolumn{1}{c}{Orig}   & \multicolumn{2}{c}{\adt{}-multi} \\
\midrule
\texttt{sw}                 & 0.510   & 0.574                       & 12.55\%                      & 0.54                       & 5.88\%                      & 0.522 & 0.546                      & 4.60\%                     \\
\texttt{id}                 & 0.584   & 0.608                       & 4.11\%                       & 0.592                      & 1.37\%                      & 0.628 & 0.604                      & -3.82\%                    \\
\texttt{et}                 & 0.470   & 0.538                       & 14.47\%                      & 0.5                        & 6.38\%                      & 0.488 & 0.538                      & 10.25\%                    \\
\texttt{ht}                 & 0.514   & 0.548                       & 6.61\%                       & 0.538                      & 4.67\%                      & 0.506 & 0.526                      & 3.95\%                     \\
\hline
Avg (4 langs)      & 0.520   & 0.567                       & 9.14\%                       & 0.542                      & 4.33\%                      & 0.536 & 0.5535                     & 3.26\%                     \\
\toprule
\multirow{2}{*}{Belebele} & \multicolumn{5}{c|}{Mistral}                                                                 & \multicolumn{3}{c}{\textbf{Llama}}                       \\ 
                    & \multicolumn{1}{c}{Orig}  & \multicolumn{2}{c}{\adt{}-multi} & \multicolumn{2}{c|}{\adt{}-\textbf{all}} & \multicolumn{1}{c}{Orig}   & \multicolumn{2}{c}{\adt{}-multi} \\
\midrule
\texttt{en}                 & 0.843 & 0.833    & -1.18\%    & 0.824    & -2.29\%    & 0.482 & 0.456    & -5.39\%    \\
\texttt{sw}                 & 0.391 & 0.440    & 12.50\%    & 0.454    & 16.08\%    & 0.262 & 0.289    & 10.31\%    \\
\texttt{id}                 & 0.647 & 0.638    & -1.38\%    & 0.636    & -1.65\%    & 0.380 & 0.346    & -8.95\%    \\
\texttt{et}                 & 0.439 & 0.538    & 22.53\%    & 0.54     & 23.03\%    & 0.312 & 0.319    & 2.24\%     \\
\texttt{ht}                 & 0.397 & 0.507    & 27.72\%    & 0.522    & 31.59\%    & 0.287 & 0.322    & 12.20\%    \\
\texttt{ko}                 & 0.666 & 0.616    & -7.52\%    & 0.644    & -3.25\%    & 0.336 & 0.354    & 5.36\%     \\
\texttt{el}                 & 0.442 & 0.566    & 27.90\%    & 0.631    & 42.70\%    & 0.301 & 0.357    & 18.60\%    \\
\texttt{ru}                 & 0.727 & 0.696    & -4.29\%    & 0.71     & -2.30\%    & 0.428 & 0.378    & -11.68\%   \\
\texttt{bg}                 & 0.674 & 0.698    & 3.47\%     & 0.694    & 2.91\%     & 0.398 & 0.392    & -1.51\%    \\
\hline
Avg (9 langs)             & 0.581 & 0.614    & 5.83\%     & 0.629    & 8.33\%     & 0.354 & 0.357    & 0.86\%     \\
\hline
\texttt{uk}                 & 0.728 & 0.693    & -4.76\%    & 0.682    & -6.32\%    & 0.398 & 0.352    & -11.50\%   \\
\texttt{kk}                 & 0.364 & 0.442    & 21.36\%    & 0.427    & 17.18\%    & 0.261 & 0.277    & 6.00\%     \\
\hline
Avg(11 langs)            & 0.574 & 0.606    & 5.50\%     & 0.615    & 7.08\%     & 0.349 & 0.349    & 0.05\%     \\
\toprule
\multirow{2}{*}{MMMLU} & \multicolumn{5}{c|}{Mistral}                                                                 & \multicolumn{3}{c}{\textbf{Llama}}                       \\ 
                    & \multicolumn{1}{c}{Orig}  & \multicolumn{2}{c}{\adt{}-multi} & \multicolumn{2}{c|}{\adt{}-\textbf{all}} & \multicolumn{1}{c}{Orig}   & \multicolumn{2}{c}{\adt{}-multi} \\
\midrule
\texttt{en}                 & 0.607 & 0.577    & -4.88\%    & 0.561    & -7.53\%    & 0.452 & 0.415    & -8.19\%    \\
\texttt{id}                 & 0.468 & 0.410    & -12.49\%   & 0.444    & -5.17\%    & 0.367 & 0.296    & -19.35\%   \\
\texttt{ru}                 & 0.500 & 0.468    & -6.39\%    & 0.468    & -6.33\%    & 0.355 & 0.34     & -4.23\%    \\
\hline
Avg (3 langs)      & 0.525 & 0.485    & -7.62\%    & 0.491    & -6.45\%    & 0.391 & 0.351    & -10.23\%   \\
\hline
\texttt{uk}                 & 0.489 & 0.462    & -5.57\%    & 0.463    & -5.34\%    & 0.346 & 0.328    & -5.20\%    \\
\hline
Avg (4 langs)      & 0.516 & 0.479    & -7.14\%    & 0.484    & -6.18\%    & 0.38  & 0.345    & -9.21\%    \\
\bottomrule
\end{tabular}
\end{table}

%% file: iclr2025_conference.bbl
\begin{thebibliography}{80}
\providecommand{\natexlab}[1]{#1}
\providecommand{\url}[1]{\texttt{#1}}
\expandafter\ifx\csname urlstyle\endcsname\relax
  \providecommand{\doi}[1]{doi: #1}\else
  \providecommand{\doi}{doi: \begingroup \urlstyle{rm}\Url}\fi

\bibitem[Aharoni et~al.(2019)Aharoni, Johnson, and Firat]{aharoni-etal-2019-massively}
Roee Aharoni, Melvin Johnson, and Orhan Firat.
\newblock Massively multilingual neural machine translation.
\newblock In Jill Burstein, Christy Doran, and Thamar Solorio (eds.), \emph{Proceedings of the 2019 Conference of the North {A}merican Chapter of the Association for Computational Linguistics: Human Language Technologies, Volume 1 (Long and Short Papers)}, pp.\  3874--3884, Minneapolis, Minnesota, June 2019. Association for Computational Linguistics.
\newblock \doi{10.18653/v1/N19-1388}.
\newblock URL \url{https://aclanthology.org/N19-1388}.

\bibitem[Ahia et~al.(2023)Ahia, Kumar, Gonen, Kasai, Mortensen, Smith, and Tsvetkov]{do-all-lang-cost-same23}
Orevaoghene Ahia, Sachin Kumar, Hila Gonen, Jungo Kasai, David Mortensen, Noah Smith, and Yulia Tsvetkov.
\newblock Do all languages cost the same? tokenization in the era of commercial language models.
\newblock In Houda Bouamor, Juan Pino, and Kalika Bali (eds.), \emph{Proceedings of the 2023 Conference on Empirical Methods in Natural Language Processing}, pp.\  9904--9923, Singapore, December 2023. Association for Computational Linguistics.
\newblock \doi{10.18653/v1/2023.emnlp-main.614}.
\newblock URL \url{https://aclanthology.org/2023.emnlp-main.614}.

\bibitem[Artetxe \& Schwenk(2019)Artetxe and Schwenk]{laser19}
Mikel Artetxe and Holger Schwenk.
\newblock {Massively Multilingual Sentence Embeddings for Zero-Shot Cross-Lingual Transfer and Beyond}.
\newblock \emph{Transactions of the Association for Computational Linguistics}, 7:\penalty0 597--610, 09 2019.
\newblock ISSN 2307-387X.
\newblock \doi{10.1162/tacl_a_00288}.
\newblock URL \url{https://doi.org/10.1162/tacl\_a\_00288}.

\bibitem[Bandarkar et~al.(2024)Bandarkar, Liang, Muller, Artetxe, Shukla, Husa, Goyal, Krishnan, Zettlemoyer, and Khabsa]{belebele23}
Lucas Bandarkar, Davis Liang, Benjamin Muller, Mikel Artetxe, Satya~Narayan Shukla, Donald Husa, Naman Goyal, Abhinandan Krishnan, Luke Zettlemoyer, and Madian Khabsa.
\newblock The belebele benchmark: a parallel reading comprehension dataset in 122 language variants.
\newblock In Lun-Wei Ku, Andre Martins, and Vivek Srikumar (eds.), \emph{Proceedings of the 62nd Annual Meeting of the Association for Computational Linguistics (Volume 1: Long Papers)}, pp.\  749--775, Bangkok, Thailand, August 2024. Association for Computational Linguistics.
\newblock \doi{10.18653/v1/2024.acl-long.44}.
\newblock URL \url{https://aclanthology.org/2024.acl-long.44}.

\bibitem[Castilho \& Knowles(2024)Castilho and Knowles]{Castilho_Knowles_2024}
Sheila Castilho and Rebecca Knowles.
\newblock A survey of context in neural machine translation and its evaluation.
\newblock \emph{Natural Language Processing}, pp.\  1–31, 2024.
\newblock \doi{10.1017/nlp.2024.7}.

\bibitem[Caswell et~al.(2020)Caswell, Breiner, van Esch, and Bapna]{caswell-etal-2020-language}
Isaac Caswell, Theresa Breiner, Daan van Esch, and Ankur Bapna.
\newblock Language {ID} in the wild: Unexpected challenges on the path to a thousand-language web text corpus.
\newblock In Donia Scott, Nuria Bel, and Chengqing Zong (eds.), \emph{Proceedings of the 28th International Conference on Computational Linguistics}, pp.\  6588--6608, Barcelona, Spain (Online), December 2020. International Committee on Computational Linguistics.
\newblock \doi{10.18653/v1/2020.coling-main.579}.
\newblock URL \url{https://aclanthology.org/2020.coling-main.579/}.

\bibitem[Chau et~al.(2020)Chau, Lin, and Smith]{chau-etal-2020-parsing}
Ethan~C. Chau, Lucy~H. Lin, and Noah~A. Smith.
\newblock Parsing with multilingual {BERT}, a small corpus, and a small treebank.
\newblock In Trevor Cohn, Yulan He, and Yang Liu (eds.), \emph{Findings of the Association for Computational Linguistics: EMNLP 2020}, pp.\  1324--1334, Online, November 2020. Association for Computational Linguistics.
\newblock \doi{10.18653/v1/2020.findings-emnlp.118}.
\newblock URL \url{https://aclanthology.org/2020.findings-emnlp.118}.

\bibitem[Chaudhary et~al.(2019)Chaudhary, Tang, Guzm{\'a}n, Schwenk, and Koehn]{chaudhary-etal-2019-low}
Vishrav Chaudhary, Yuqing Tang, Francisco Guzm{\'a}n, Holger Schwenk, and Philipp Koehn.
\newblock Low-resource corpus filtering using multilingual sentence embeddings.
\newblock In Ond{\v{r}}ej Bojar, Rajen Chatterjee, Christian Federmann, Mark Fishel, Yvette Graham, Barry Haddow, Matthias Huck, Antonio~Jimeno Yepes, Philipp Koehn, Andr{\'e} Martins, Christof Monz, Matteo Negri, Aur{\'e}lie N{\'e}v{\'e}ol, Mariana Neves, Matt Post, Marco Turchi, and Karin Verspoor (eds.), \emph{Proceedings of the Fourth Conference on Machine Translation (Volume 3: Shared Task Papers, Day 2)}, pp.\  261--266, Florence, Italy, August 2019. Association for Computational Linguistics.
\newblock \doi{10.18653/v1/W19-5435}.
\newblock URL \url{https://aclanthology.org/W19-5435/}.

\bibitem[Chung et~al.(2020)Chung, Garrette, Tan, and Riesa]{chung-etal-2020-improving}
Hyung~Won Chung, Dan Garrette, Kiat~Chuan Tan, and Jason Riesa.
\newblock Improving multilingual models with language-clustered vocabularies.
\newblock In Bonnie Webber, Trevor Cohn, Yulan He, and Yang Liu (eds.), \emph{Proceedings of the 2020 Conference on Empirical Methods in Natural Language Processing (EMNLP)}, pp.\  4536--4546, Online, November 2020. Association for Computational Linguistics.
\newblock \doi{10.18653/v1/2020.emnlp-main.367}.
\newblock URL \url{https://aclanthology.org/2020.emnlp-main.367}.

\bibitem[Conneau et~al.(2018)Conneau, Rinott, Lample, Williams, Bowman, Schwenk, and Stoyanov]{xnli}
Alexis Conneau, Ruty Rinott, Guillaume Lample, Adina Williams, Samuel Bowman, Holger Schwenk, and Veselin Stoyanov.
\newblock {XNLI}: Evaluating cross-lingual sentence representations.
\newblock In Ellen Riloff, David Chiang, Julia Hockenmaier, and Jun{'}ichi Tsujii (eds.), \emph{Proceedings of the 2018 Conference on Empirical Methods in Natural Language Processing}, pp.\  2475--2485, Brussels, Belgium, October-November 2018. Association for Computational Linguistics.
\newblock \doi{10.18653/v1/D18-1269}.
\newblock URL \url{https://aclanthology.org/D18-1269}.

\bibitem[Deutsch et~al.(2021)Deutsch, Dror, and Roth]{deutsch-etal-2021-statistical}
Daniel Deutsch, Rotem Dror, and Dan Roth.
\newblock A statistical analysis of summarization evaluation metrics using resampling methods.
\newblock \emph{Transactions of the Association for Computational Linguistics}, 9:\penalty0 1132--1146, 2021.
\newblock \doi{10.1162/tacl_a_00417}.
\newblock URL \url{https://aclanthology.org/2021.tacl-1.67/}.

\bibitem[Deutsch et~al.(2023)Deutsch, Juraska, Finkelstein, and Freitag]{deutsch-etal-2023-training}
Daniel Deutsch, Juraj Juraska, Mara Finkelstein, and Markus Freitag.
\newblock Training and meta-evaluating machine translation evaluation metrics at the paragraph level.
\newblock In Philipp Koehn, Barry Haddow, Tom Kocmi, and Christof Monz (eds.), \emph{Proceedings of the Eighth Conference on Machine Translation}, pp.\  996--1013, Singapore, December 2023. Association for Computational Linguistics.
\newblock \doi{10.18653/v1/2023.wmt-1.96}.
\newblock URL \url{https://aclanthology.org/2023.wmt-1.96/}.

\bibitem[Dobler \& de~Melo(2023)Dobler and de~Melo]{focus23}
Konstantin Dobler and Gerard de~Melo.
\newblock {FOCUS}: Effective embedding initialization for monolingual specialization of multilingual models.
\newblock In Houda Bouamor, Juan Pino, and Kalika Bali (eds.), \emph{Proceedings of the 2023 Conference on Empirical Methods in Natural Language Processing}, pp.\  13440--13454, Singapore, December 2023. Association for Computational Linguistics.
\newblock \doi{10.18653/v1/2023.emnlp-main.829}.
\newblock URL \url{https://aclanthology.org/2023.emnlp-main.829}.

\bibitem[Dodge et~al.(2021)Dodge, Sap, Marasovi{\'c}, Agnew, Ilharco, Groeneveld, Mitchell, and Gardner]{dodge-etal-2021-documenting}
Jesse Dodge, Maarten Sap, Ana Marasovi{\'c}, William Agnew, Gabriel Ilharco, Dirk Groeneveld, Margaret Mitchell, and Matt Gardner.
\newblock Documenting large webtext corpora: A case study on the colossal clean crawled corpus.
\newblock In Marie-Francine Moens, Xuanjing Huang, Lucia Specia, and Scott Wen-tau Yih (eds.), \emph{Proceedings of the 2021 Conference on Empirical Methods in Natural Language Processing}, pp.\  1286--1305, Online and Punta Cana, Dominican Republic, November 2021. Association for Computational Linguistics.
\newblock \doi{10.18653/v1/2021.emnlp-main.98}.
\newblock URL \url{https://aclanthology.org/2021.emnlp-main.98/}.

\bibitem[Downey et~al.(2023)Downey, Blevins, Goldfine, and Steinert-Threlkeld]{downey-etal-2023-embedding}
C.m. Downey, Terra Blevins, Nora Goldfine, and Shane Steinert-Threlkeld.
\newblock Embedding structure matters: Comparing methods to adapt multilingual vocabularies to new languages.
\newblock In Duygu Ataman (ed.), \emph{Proceedings of the 3rd Workshop on Multi-lingual Representation Learning (MRL)}, pp.\  268--281, Singapore, December 2023. Association for Computational Linguistics.
\newblock \doi{10.18653/v1/2023.mrl-1.20}.
\newblock URL \url{https://aclanthology.org/2023.mrl-1.20}.

\bibitem[Freitag et~al.(2023)Freitag, Mathur, Lo, Avramidis, Rei, Thompson, Kocmi, Blain, Deutsch, Stewart, Zerva, Castilho, Lavie, and Foster]{freitag-etal-2023-results}
Markus Freitag, Nitika Mathur, Chi-kiu Lo, Eleftherios Avramidis, Ricardo Rei, Brian Thompson, Tom Kocmi, Frederic Blain, Daniel Deutsch, Craig Stewart, Chrysoula Zerva, Sheila Castilho, Alon Lavie, and George Foster.
\newblock Results of {WMT}23 metrics shared task: Metrics might be guilty but references are not innocent.
\newblock In Philipp Koehn, Barry Haddow, Tom Kocmi, and Christof Monz (eds.), \emph{Proceedings of the Eighth Conference on Machine Translation}, pp.\  578--628, Singapore, December 2023. Association for Computational Linguistics.
\newblock \doi{10.18653/v1/2023.wmt-1.51}.
\newblock URL \url{https://aclanthology.org/2023.wmt-1.51/}.

\bibitem[Freitag et~al.(2024)Freitag, Mathur, Deutsch, Lo, Avramidis, Rei, Thompson, Blain, Kocmi, Wang, Adelani, Buchicchio, Zerva, and Lavie]{freitag-etal-2024-llms}
Markus Freitag, Nitika Mathur, Daniel Deutsch, Chi-Kiu Lo, Eleftherios Avramidis, Ricardo Rei, Brian Thompson, Frederic Blain, Tom Kocmi, Jiayi Wang, David~Ifeoluwa Adelani, Marianna Buchicchio, Chrysoula Zerva, and Alon Lavie.
\newblock Are {LLM}s breaking {MT} metrics? results of the {WMT}24 metrics shared task.
\newblock In Barry Haddow, Tom Kocmi, Philipp Koehn, and Christof Monz (eds.), \emph{Proceedings of the Ninth Conference on Machine Translation}, pp.\  47--81, Miami, Florida, USA, November 2024. Association for Computational Linguistics.
\newblock \doi{10.18653/v1/2024.wmt-1.2}.
\newblock URL \url{https://aclanthology.org/2024.wmt-1.2/}.

\bibitem[Gao et~al.(2024)Gao, Tow, Abbasi, Biderman, Black, DiPofi, Foster, Golding, Hsu, Le~Noac'h, Li, McDonell, Muennighoff, Ociepa, Phang, Reynolds, Schoelkopf, Skowron, Sutawika, Tang, Thite, Wang, Wang, and Zou]{lmeval24}
Leo Gao, Jonathan Tow, Baber Abbasi, Stella Biderman, Sid Black, Anthony DiPofi, Charles Foster, Laurence Golding, Jeffrey Hsu, Alain Le~Noac'h, Haonan Li, Kyle McDonell, Niklas Muennighoff, Chris Ociepa, Jason Phang, Laria Reynolds, Hailey Schoelkopf, Aviya Skowron, Lintang Sutawika, Eric Tang, Anish Thite, Ben Wang, Kevin Wang, and Andy Zou.
\newblock A framework for few-shot language model evaluation, 07 2024.
\newblock URL \url{https://zenodo.org/records/12608602}.

\bibitem[Gee et~al.(2022)Gee, Zugarini, Rigutini, and Torroni]{fvt22}
Leonidas Gee, Andrea Zugarini, Leonardo Rigutini, and Paolo Torroni.
\newblock Fast vocabulary transfer for language model compression.
\newblock In Yunyao Li and Angeliki Lazaridou (eds.), \emph{Proceedings of the 2022 Conference on Empirical Methods in Natural Language Processing: Industry Track}, pp.\  409--416, Abu Dhabi, UAE, December 2022. Association for Computational Linguistics.
\newblock \doi{10.18653/v1/2022.emnlp-industry.41}.
\newblock URL \url{https://aclanthology.org/2022.emnlp-industry.41}.

\bibitem[Gogoulou et~al.(2022)Gogoulou, Ekgren, Isbister, and Sahlgren]{gogoulou-etal-2022-cross}
Evangelia Gogoulou, Ariel Ekgren, Tim Isbister, and Magnus Sahlgren.
\newblock Cross-lingual transfer of monolingual models.
\newblock In Nicoletta Calzolari, Fr{\'e}d{\'e}ric B{\'e}chet, Philippe Blache, Khalid Choukri, Christopher Cieri, Thierry Declerck, Sara Goggi, Hitoshi Isahara, Bente Maegaard, Joseph Mariani, H{\'e}l{\`e}ne Mazo, Jan Odijk, and Stelios Piperidis (eds.), \emph{Proceedings of the Thirteenth Language Resources and Evaluation Conference}, pp.\  948--955, Marseille, France, June 2022. European Language Resources Association.
\newblock URL \url{https://aclanthology.org/2022.lrec-1.100}.

\bibitem[Goyal et~al.(2022)Goyal, Gao, Chaudhary, Chen, Wenzek, Ju, Krishnan, Ranzato, Guzmán, and Fan]{flores101}
Naman Goyal, Cynthia Gao, Vishrav Chaudhary, Peng-Jen Chen, Guillaume Wenzek, Da~Ju, Sanjana Krishnan, Marc’Aurelio Ranzato, Francisco Guzmán, and Angela Fan.
\newblock {The Flores-101 Evaluation Benchmark for Low-Resource and Multilingual Machine Translation}.
\newblock \emph{Transactions of the Association for Computational Linguistics}, 10:\penalty0 522--538, 05 2022.
\newblock ISSN 2307-387X.
\newblock \doi{10.1162/tacl_a_00474}.
\newblock URL \url{https://doi.org/10.1162/tacl\_a\_00474}.

\bibitem[Guerreiro et~al.(2023)Guerreiro, Rei, van Stigt, Coheur, Colombo, and Martins]{xcomet23}
Nuno~M. Guerreiro, Ricardo Rei, Daan van Stigt, Luisa Coheur, Pierre Colombo, and André F.~T. Martins.
\newblock xcomet: Transparent machine translation evaluation through fine-grained error detection, 2023.
\newblock URL \url{https://arxiv.org/abs/2310.10482}.

\bibitem[Heffernan et~al.(2022)Heffernan, {\c{C}}elebi, and Schwenk]{heffernan-etal-2022-bitext}
Kevin Heffernan, Onur {\c{C}}elebi, and Holger Schwenk.
\newblock Bitext mining using distilled sentence representations for low-resource languages.
\newblock In Yoav Goldberg, Zornitsa Kozareva, and Yue Zhang (eds.), \emph{Findings of the Association for Computational Linguistics: EMNLP 2022}, pp.\  2101--2112, Abu Dhabi, United Arab Emirates, December 2022. Association for Computational Linguistics.
\newblock \doi{10.18653/v1/2022.findings-emnlp.154}.
\newblock URL \url{https://aclanthology.org/2022.findings-emnlp.154}.

\bibitem[Hendrycks et~al.(2021)Hendrycks, Burns, Basart, Zou, Mazeika, Song, and Steinhardt]{mmlu21}
Dan Hendrycks, Collin Burns, Steven Basart, Andy Zou, Mantas Mazeika, Dawn Song, and Jacob Steinhardt.
\newblock Measuring massive multitask language understanding.
\newblock In \emph{International Conference on Learning Representations}, 2021.
\newblock URL \url{https://openreview.net/forum?id=d7KBjmI3GmQ}.

\bibitem[Houlsby et~al.(2019)Houlsby, Giurgiu, Jastrzebski, Morrone, De~Laroussilhe, Gesmundo, Attariyan, and Gelly]{pmlr-v97-houlsby19a}
Neil Houlsby, Andrei Giurgiu, Stanislaw Jastrzebski, Bruna Morrone, Quentin De~Laroussilhe, Andrea Gesmundo, Mona Attariyan, and Sylvain Gelly.
\newblock Parameter-efficient transfer learning for {NLP}.
\newblock In Kamalika Chaudhuri and Ruslan Salakhutdinov (eds.), \emph{Proceedings of the 36th International Conference on Machine Learning}, volume~97 of \emph{Proceedings of Machine Learning Research}, pp.\  2790--2799. PMLR, 09--15 Jun 2019.
\newblock URL \url{https://proceedings.mlr.press/v97/houlsby19a.html}.

\bibitem[Hu et~al.(2022)Hu, yelong shen, Wallis, Allen-Zhu, Li, Wang, Wang, and Chen]{hu2022lora}
Edward~J Hu, yelong shen, Phillip Wallis, Zeyuan Allen-Zhu, Yuanzhi Li, Shean Wang, Lu~Wang, and Weizhu Chen.
\newblock Lo{RA}: Low-rank adaptation of large language models.
\newblock In \emph{International Conference on Learning Representations}, 2022.
\newblock URL \url{https://openreview.net/forum?id=nZeVKeeFYf9}.

\bibitem[Jiang et~al.(2023)Jiang, Sablayrolles, Mensch, Bamford, Chaplot, de~las Casas, Bressand, Lengyel, Lample, Saulnier, Lavaud, Lachaux, Stock, Scao, Lavril, Wang, Lacroix, and Sayed]{mistral7b}
Albert~Q. Jiang, Alexandre Sablayrolles, Arthur Mensch, Chris Bamford, Devendra~Singh Chaplot, Diego de~las Casas, Florian Bressand, Gianna Lengyel, Guillaume Lample, Lucile Saulnier, Lélio~Renard Lavaud, Marie-Anne Lachaux, Pierre Stock, Teven~Le Scao, Thibaut Lavril, Thomas Wang, Timothée Lacroix, and William~El Sayed.
\newblock Mistral 7b, 2023.
\newblock URL \url{https://arxiv.org/abs/2310.06825}.

\bibitem[Joshi et~al.(2020)Joshi, Santy, Budhiraja, Bali, and Choudhury]{joshi-etal-2020-state}
Pratik Joshi, Sebastin Santy, Amar Budhiraja, Kalika Bali, and Monojit Choudhury.
\newblock The state and fate of linguistic diversity and inclusion in the {NLP} world.
\newblock In Dan Jurafsky, Joyce Chai, Natalie Schluter, and Joel Tetreault (eds.), \emph{Proceedings of the 58th Annual Meeting of the Association for Computational Linguistics}, pp.\  6282--6293, Online, July 2020. Association for Computational Linguistics.
\newblock \doi{10.18653/v1/2020.acl-main.560}.
\newblock URL \url{https://aclanthology.org/2020.acl-main.560}.

\bibitem[Karpinska \& Iyyer(2023)Karpinska and Iyyer]{karpinska-iyyer-2023-large}
Marzena Karpinska and Mohit Iyyer.
\newblock Large language models effectively leverage document-level context for literary translation, but critical errors persist.
\newblock In Philipp Koehn, Barry Haddow, Tom Kocmi, and Christof Monz (eds.), \emph{Proceedings of the Eighth Conference on Machine Translation}, pp.\  419--451, Singapore, December 2023. Association for Computational Linguistics.
\newblock \doi{10.18653/v1/2023.wmt-1.41}.
\newblock URL \url{https://aclanthology.org/2023.wmt-1.41/}.

\bibitem[Khayrallah \& Koehn(2018)Khayrallah and Koehn]{khayrallah-koehn-2018-impact}
Huda Khayrallah and Philipp Koehn.
\newblock On the impact of various types of noise on neural machine translation.
\newblock In Alexandra Birch, Andrew Finch, Thang Luong, Graham Neubig, and Yusuke Oda (eds.), \emph{Proceedings of the 2nd Workshop on Neural Machine Translation and Generation}, pp.\  74--83, Melbourne, Australia, July 2018. Association for Computational Linguistics.
\newblock \doi{10.18653/v1/W18-2709}.
\newblock URL \url{https://aclanthology.org/W18-2709/}.

\bibitem[Khayrallah et~al.(2018)Khayrallah, Thompson, Duh, and Koehn]{khayrallah-etal-2018-regularized}
Huda Khayrallah, Brian Thompson, Kevin Duh, and Philipp Koehn.
\newblock Regularized training objective for continued training for domain adaptation in neural machine translation.
\newblock In Alexandra Birch, Andrew Finch, Thang Luong, Graham Neubig, and Yusuke Oda (eds.), \emph{Proceedings of the 2nd Workshop on Neural Machine Translation and Generation}, pp.\  36--44, Melbourne, Australia, July 2018. Association for Computational Linguistics.
\newblock \doi{10.18653/v1/W18-2705}.
\newblock URL \url{https://aclanthology.org/W18-2705/}.

\bibitem[Kirkpatrick et~al.(2017)Kirkpatrick, Pascanu, Rabinowitz, Veness, Desjardins, Rusu, Milan, Quan, Ramalho, Grabska-Barwinska, Hassabis, Clopath, Kumaran, and Hadsell]{doi:10.1073/pnas.1611835114}
James Kirkpatrick, Razvan Pascanu, Neil Rabinowitz, Joel Veness, Guillaume Desjardins, Andrei~A. Rusu, Kieran Milan, John Quan, Tiago Ramalho, Agnieszka Grabska-Barwinska, Demis Hassabis, Claudia Clopath, Dharshan Kumaran, and Raia Hadsell.
\newblock Overcoming catastrophic forgetting in neural networks.
\newblock \emph{Proceedings of the National Academy of Sciences}, 114\penalty0 (13):\penalty0 3521--3526, 2017.
\newblock \doi{10.1073/pnas.1611835114}.
\newblock URL \url{https://www.pnas.org/doi/abs/10.1073/pnas.1611835114}.

\bibitem[Koehn(2004)]{koehn-2004-statistical}
Philipp Koehn.
\newblock Statistical significance tests for machine translation evaluation.
\newblock In Dekang Lin and Dekai Wu (eds.), \emph{Proceedings of the 2004 Conference on Empirical Methods in Natural Language Processing}, pp.\  388--395, Barcelona, Spain, July 2004. Association for Computational Linguistics.
\newblock URL \url{https://aclanthology.org/W04-3250/}.

\bibitem[Koehn et~al.(2020)Koehn, Chaudhary, El-Kishky, Goyal, Chen, and Guzm{\'a}n]{koehn-etal-2020-findings}
Philipp Koehn, Vishrav Chaudhary, Ahmed El-Kishky, Naman Goyal, Peng-Jen Chen, and Francisco Guzm{\'a}n.
\newblock Findings of the {WMT} 2020 shared task on parallel corpus filtering and alignment.
\newblock In Lo{\"i}c Barrault, Ond{\v{r}}ej Bojar, Fethi Bougares, Rajen Chatterjee, Marta~R. Costa-juss{\`a}, Christian Federmann, Mark Fishel, Alexander Fraser, Yvette Graham, Paco Guzman, Barry Haddow, Matthias Huck, Antonio~Jimeno Yepes, Philipp Koehn, Andr{\'e} Martins, Makoto Morishita, Christof Monz, Masaaki Nagata, Toshiaki Nakazawa, and Matteo Negri (eds.), \emph{Proceedings of the Fifth Conference on Machine Translation}, pp.\  726--742, Online, November 2020. Association for Computational Linguistics.
\newblock URL \url{https://aclanthology.org/2020.wmt-1.78/}.

\bibitem[Kreutzer et~al.(2022)Kreutzer, Caswell, Wang, Wahab, van Esch, Ulzii-Orshikh, Tapo, Subramani, Sokolov, Sikasote, Setyawan, Sarin, Samb, Sagot, Rivera, Rios, Papadimitriou, Osei, Suarez, Orife, Ogue\~ji, Rubungo, Nguyen, Müller, Müller, Muhammad, Muhammad, Mnyakeni, Mirzakhalov, Matangir\~a, Leong, Lawson, Kudugunta, Jernite, Jenny, Firat, Dossou, Dlamini, de~Silva, Çabuk Ballı\, Biderman, Battisti, Baruwa, Bapna, Baljekar, Azime, Awokoya, Ataman, Ahia, Ahia, Agrawal, and Adeyemi]{10.1162/tacl_a_00447}
Julia Kreutzer, Isaac Caswell, Lisa Wang, Ahsan Wahab, Daan van Esch, Nasanbayar Ulzii-Orshikh, Allahsera Tapo, Nishant Subramani, Artem Sokolov, Clayton\~e Sikasote, Monang Setyawan, Supheakmungkol Sarin, Sokhar Samb, Benoît Sagot, Clara Rivera, Annette Rios, Isabel Papadimitriou, Salomey Osei, Pedro~Ortiz Suarez, Iroro Orife, Kelechi Ogue\~ji, Andre~Niyongabo Rubungo, Toan~Q. Nguyen, Mathias Müller, André Müller, Shamsuddeen~Hassan Muhammad, Nanda Muhammad, Ayanda Mnyakeni, Jamshidbek Mirzakhalov, Tapiwanashe Matangir\~a, Colin Leong, Nze Lawson, Sneha Kudugunta, Yacine Jernite, Mathias Jenny, Orhan Firat, Bonaventure F.~P. Dossou, Sakhile Dlamini, Nisansa de~Silva, Sakine Çabuk Ballı\, Stella Biderman, Alessia Battisti, Ahmed Baruwa, Ankur Bapna, Pallavi Baljekar, Israel~Abebe Azime, Ayodele Awokoya, Duygu Ataman, Orevaoghene Ahia, Oghenefeg\~o Ahia, Sweta Agrawal, and Mofetoluwa Adeyemi.
\newblock Quality at a glance: An audit of web-crawled multilingual datasets.
\newblock \emph{Transactions of the Association for Computational Linguistics}, 10:\penalty0 50--72, 01 2022.
\newblock ISSN 2307-387X.
\newblock \doi{10.1162/tacl_a_00447}.
\newblock URL \url{https://doi.org/10.1162/tacl\_a\_00447}.

\bibitem[Kudo \& Richardson(2018)Kudo and Richardson]{sentencepiece18}
Taku Kudo and John Richardson.
\newblock {S}entence{P}iece: A simple and language independent subword tokenizer and detokenizer for neural text processing.
\newblock In Eduardo Blanco and Wei Lu (eds.), \emph{Proceedings of the 2018 Conference on Empirical Methods in Natural Language Processing: System Demonstrations}, pp.\  66--71, Brussels, Belgium, November 2018. Association for Computational Linguistics.
\newblock \doi{10.18653/v1/D18-2012}.
\newblock URL \url{https://aclanthology.org/D18-2012}.

\bibitem[Kudugunta et~al.(2023)Kudugunta, Caswell, Zhang, Garcia, Choquette-Choo, Lee, Xin, Kusupati, Stella, Bapna, and Firat]{madlad23}
Sneha Kudugunta, Isaac Caswell, Biao Zhang, Xavier Garcia, Christopher~A. Choquette-Choo, Katherine Lee, Derrick Xin, Aditya Kusupati, Romi Stella, Ankur Bapna, and Orhan Firat.
\newblock Madlad-400: A multilingual and document-level large audited dataset, 2023.
\newblock URL \url{https://arxiv.org/abs/2309.04662}.

\bibitem[Lai et~al.(2023)Lai, Nguyen, Ngo, Nguyen, Dernoncourt, Rossi, and Nguyen]{okapi23}
Viet Lai, Chien Nguyen, Nghia Ngo, Thuat Nguyen, Franck Dernoncourt, Ryan Rossi, and Thien Nguyen.
\newblock Okapi: Instruction-tuned large language models in multiple languages with reinforcement learning from human feedback.
\newblock In Yansong Feng and Els Lefever (eds.), \emph{Proceedings of the 2023 Conference on Empirical Methods in Natural Language Processing: System Demonstrations}, pp.\  318--327, Singapore, December 2023. Association for Computational Linguistics.
\newblock \doi{10.18653/v1/2023.emnlp-demo.28}.
\newblock URL \url{https://aclanthology.org/2023.emnlp-demo.28}.

\bibitem[Liu et~al.(2024)Liu, Lin, Wang, and Schuetze]{ofa24}
Yihong Liu, Peiqin Lin, Mingyang Wang, and Hinrich Schuetze.
\newblock {OFA}: A framework of initializing unseen subword embeddings for efficient large-scale multilingual continued pretraining.
\newblock In Kevin Duh, Helena Gomez, and Steven Bethard (eds.), \emph{Findings of the Association for Computational Linguistics: NAACL 2024}, pp.\  1067--1097, Mexico City, Mexico, June 2024. Association for Computational Linguistics.
\newblock \doi{10.18653/v1/2024.findings-naacl.68}.
\newblock URL \url{https://aclanthology.org/2024.findings-naacl.68}.

\bibitem[Lo(2019)]{lo-2019-yisi}
Chi-kiu Lo.
\newblock {Y}i{S}i - a unified semantic {MT} quality evaluation and estimation metric for languages with different levels of available resources.
\newblock In Ond{\v{r}}ej Bojar, Rajen Chatterjee, Christian Federmann, Mark Fishel, Yvette Graham, Barry Haddow, Matthias Huck, Antonio~Jimeno Yepes, Philipp Koehn, Andr{\'e} Martins, Christof Monz, Matteo Negri, Aur{\'e}lie N{\'e}v{\'e}ol, Mariana Neves, Matt Post, Marco Turchi, and Karin Verspoor (eds.), \emph{Proceedings of the Fourth Conference on Machine Translation (Volume 2: Shared Task Papers, Day 1)}, pp.\  507--513, Florence, Italy, August 2019. Association for Computational Linguistics.
\newblock \doi{10.18653/v1/W19-5358}.
\newblock URL \url{https://aclanthology.org/W19-5358/}.

\bibitem[Lo \& Larkin(2020)Lo and Larkin]{lo-larkin-2020-machine}
Chi-kiu Lo and Samuel Larkin.
\newblock Machine translation reference-less evaluation using {Y}i{S}i-2 with bilingual mappings of massive multilingual language model.
\newblock In Lo{\"i}c Barrault, Ond{\v{r}}ej Bojar, Fethi Bougares, Rajen Chatterjee, Marta~R. Costa-juss{\`a}, Christian Federmann, Mark Fishel, Alexander Fraser, Yvette Graham, Paco Guzman, Barry Haddow, Matthias Huck, Antonio~Jimeno Yepes, Philipp Koehn, Andr{\'e} Martins, Makoto Morishita, Christof Monz, Masaaki Nagata, Toshiaki Nakazawa, and Matteo Negri (eds.), \emph{Proceedings of the Fifth Conference on Machine Translation}, pp.\  903--910, Online, November 2020. Association for Computational Linguistics.
\newblock URL \url{https://aclanthology.org/2020.wmt-1.100/}.

\bibitem[Lo et~al.(2023)Lo, Knowles, and Goutte]{lo-etal-2023-beyond}
Chi-kiu Lo, Rebecca Knowles, and Cyril Goutte.
\newblock Beyond correlation: Making sense of the score differences of new {MT} evaluation metrics.
\newblock In Masao Utiyama and Rui Wang (eds.), \emph{Proceedings of Machine Translation Summit XIX, Vol. 1: Research Track}, pp.\  186--199, Macau SAR, China, September 2023. Asia-Pacific Association for Machine Translation.
\newblock URL \url{https://aclanthology.org/2023.mtsummit-research.16/}.

\bibitem[Miceli~Barone et~al.(2017)Miceli~Barone, Haddow, Germann, and Sennrich]{miceli-barone-etal-2017-regularization}
Antonio~Valerio Miceli~Barone, Barry Haddow, Ulrich Germann, and Rico Sennrich.
\newblock Regularization techniques for fine-tuning in neural machine translation.
\newblock In Martha Palmer, Rebecca Hwa, and Sebastian Riedel (eds.), \emph{Proceedings of the 2017 Conference on Empirical Methods in Natural Language Processing}, pp.\  1489--1494, Copenhagen, Denmark, September 2017. Association for Computational Linguistics.
\newblock \doi{10.18653/v1/D17-1156}.
\newblock URL \url{https://aclanthology.org/D17-1156/}.

\bibitem[Minixhofer et~al.(2022)Minixhofer, Paischer, and Rekabsaz]{wechsel22}
Benjamin Minixhofer, Fabian Paischer, and Navid Rekabsaz.
\newblock {WECHSEL}: Effective initialization of subword embeddings for cross-lingual transfer of monolingual language models.
\newblock In Marine Carpuat, Marie-Catherine de~Marneffe, and Ivan~Vladimir Meza~Ruiz (eds.), \emph{Proceedings of the 2022 Conference of the North American Chapter of the Association for Computational Linguistics: Human Language Technologies}, pp.\  3992--4006, Seattle, United States, July 2022. Association for Computational Linguistics.
\newblock \doi{10.18653/v1/2022.naacl-main.293}.
\newblock URL \url{https://aclanthology.org/2022.naacl-main.293}.

\bibitem[Minixhofer et~al.(2024)Minixhofer, Ponti, and Vulić]{zett24}
Benjamin Minixhofer, Edoardo~Maria Ponti, and Ivan Vulić.
\newblock Zero-shot tokenizer transfer, 2024.
\newblock URL \url{https://arxiv.org/abs/2405.07883}.

\bibitem[Mosin et~al.(2023)Mosin, Samenko, Kozlovskii, Tikhonov, and Yamshchikov]{vipi22}
Vladislav Mosin, Igor Samenko, Borislav Kozlovskii, Alexey Tikhonov, and Ivan~P. Yamshchikov.
\newblock Fine-tuning transformers: Vocabulary transfer.
\newblock \emph{Artificial Intelligence}, 317:\penalty0 103860, 2023.
\newblock ISSN 0004-3702.
\newblock \doi{https://doi.org/10.1016/j.artint.2023.103860}.
\newblock URL \url{https://www.sciencedirect.com/science/article/pii/S0004370223000061}.

\bibitem[Mundra et~al.(2024)Mundra, Kishore, Dabre, Puduppully, Kunchukuttan, and Khapra]{cw2v24}
Nandini Mundra, Aditya~Nanda Kishore, Raj Dabre, Ratish Puduppully, Anoop Kunchukuttan, and Mitesh~M. Khapra.
\newblock An empirical comparison of vocabulary expansion and initialization approaches for language models, 2024.
\newblock URL \url{https://arxiv.org/abs/2407.05841}.

\bibitem[{NLLB Team} et~al.(2022){NLLB Team}, Costa-jussà, Cross, Çelebi, Elbayad, Heafield, Heffernan, Kalbassi, Lam, Licht, Maillard, Sun, Wang, Wenzek, Youngblood, Akula, Barrault, Mejia-Gonzalez, Hansanti, Hoffman, Jarrett, Sadagopan, Rowe, Spruit, Tran, Andrews, Ayan, Bhosale, Edunov, Fan, Gao, Goswami, Guzmán, Koehn, Mourachko, Ropers, Saleem, Schwenk, and Wang]{flores22}
{NLLB Team}, Marta~R. Costa-jussà, James Cross, Onur Çelebi, Maha Elbayad, Kenneth Heafield, Kevin Heffernan, Elahe Kalbassi, Janice Lam, Daniel Licht, Jean Maillard, Anna Sun, Skyler Wang, Guillaume Wenzek, Al~Youngblood, Bapi Akula, Loic Barrault, Gabriel Mejia-Gonzalez, Prangthip Hansanti, John Hoffman, Semarley Jarrett, Kaushik~Ram Sadagopan, Dirk Rowe, Shannon Spruit, Chau Tran, Pierre Andrews, Necip~Fazil Ayan, Shruti Bhosale, Sergey Edunov, Angela Fan, Cynthia Gao, Vedanuj Goswami, Francisco Guzmán, Philipp Koehn, Alexandre Mourachko, Christophe Ropers, Safiyyah Saleem, Holger Schwenk, and Jeff Wang.
\newblock No language left behind: Scaling human-centered machine translation.
\newblock 2022.
\newblock URL \url{https://arxiv.org/abs/2207.04672}.

\bibitem[Ostendorff \& Rehm(2023)Ostendorff and Rehm]{clp23}
Malte Ostendorff and Georg Rehm.
\newblock Efficient language model training through cross-lingual and progressive transfer learning, 2023.
\newblock URL \url{https://arxiv.org/abs/2301.09626}.

\bibitem[Papineni et~al.(2002)Papineni, Roukos, Ward, and Zhu]{papineni-etal-2002-bleu}
Kishore Papineni, Salim Roukos, Todd Ward, and Wei-Jing Zhu.
\newblock {B}leu: a method for automatic evaluation of machine translation.
\newblock In Pierre Isabelle, Eugene Charniak, and Dekang Lin (eds.), \emph{Proceedings of the 40th Annual Meeting of the Association for Computational Linguistics}, pp.\  311--318, Philadelphia, Pennsylvania, USA, July 2002. Association for Computational Linguistics.
\newblock \doi{10.3115/1073083.1073135}.
\newblock URL \url{https://aclanthology.org/P02-1040/}.

\bibitem[Petrov et~al.(2023)Petrov, Malfa, Torr, and Bibi]{token-unfairness23}
Aleksandar Petrov, Emanuele~La Malfa, Philip Torr, and Adel Bibi.
\newblock Language model tokenizers introduce unfairness between languages.
\newblock In \emph{Thirty-seventh Conference on Neural Information Processing Systems}, 2023.
\newblock URL \url{https://openreview.net/forum?id=78yDLKi95p}.

\bibitem[Pfeiffer et~al.(2020)Pfeiffer, Vuli{\'c}, Gurevych, and Ruder]{madx20}
Jonas Pfeiffer, Ivan Vuli{\'c}, Iryna Gurevych, and Sebastian Ruder.
\newblock {MAD-X}: {A}n {A}dapter-{B}ased {F}ramework for {M}ulti-{T}ask {C}ross-{L}ingual {T}ransfer.
\newblock In Bonnie Webber, Trevor Cohn, Yulan He, and Yang Liu (eds.), \emph{Proceedings of the 2020 Conference on Empirical Methods in Natural Language Processing (EMNLP)}, pp.\  7654--7673, Online, November 2020. Association for Computational Linguistics.
\newblock \doi{10.18653/v1/2020.emnlp-main.617}.
\newblock URL \url{https://aclanthology.org/2020.emnlp-main.617}.

\bibitem[Ponti et~al.(2020)Ponti, Glava{\v{s}}, Majewska, Liu, Vuli{\'c}, and Korhonen]{xcopa20}
Edoardo~Maria Ponti, Goran Glava{\v{s}}, Olga Majewska, Qianchu Liu, Ivan Vuli{\'c}, and Anna Korhonen.
\newblock {XCOPA}: A multilingual dataset for causal commonsense reasoning.
\newblock In Bonnie Webber, Trevor Cohn, Yulan He, and Yang Liu (eds.), \emph{Proceedings of the 2020 Conference on Empirical Methods in Natural Language Processing (EMNLP)}, pp.\  2362--2376, Online, November 2020. Association for Computational Linguistics.
\newblock \doi{10.18653/v1/2020.emnlp-main.185}.
\newblock URL \url{https://aclanthology.org/2020.emnlp-main.185}.

\bibitem[Popovi{\'c}(2015)]{popovic-2015-chrf}
Maja Popovi{\'c}.
\newblock chr{F}: character n-gram {F}-score for automatic {MT} evaluation.
\newblock In Ond{\v{r}}ej Bojar, Rajan Chatterjee, Christian Federmann, Barry Haddow, Chris Hokamp, Matthias Huck, Varvara Logacheva, and Pavel Pecina (eds.), \emph{Proceedings of the Tenth Workshop on Statistical Machine Translation}, pp.\  392--395, Lisbon, Portugal, September 2015. Association for Computational Linguistics.
\newblock \doi{10.18653/v1/W15-3049}.
\newblock URL \url{https://aclanthology.org/W15-3049/}.

\bibitem[Popovi{\'c}(2017)]{popovic-2017-chrf}
Maja Popovi{\'c}.
\newblock chr{F}++: words helping character n-grams.
\newblock In Ond{\v{r}}ej Bojar, Christian Buck, Rajen Chatterjee, Christian Federmann, Yvette Graham, Barry Haddow, Matthias Huck, Antonio~Jimeno Yepes, Philipp Koehn, and Julia Kreutzer (eds.), \emph{Proceedings of the Second Conference on Machine Translation}, pp.\  612--618, Copenhagen, Denmark, September 2017. Association for Computational Linguistics.
\newblock \doi{10.18653/v1/W17-4770}.
\newblock URL \url{https://aclanthology.org/W17-4770/}.

\bibitem[Raunak et~al.(2024)Raunak, Kocmi, and Post]{raunak-etal-2024-slide}
Vikas Raunak, Tom Kocmi, and Matt Post.
\newblock {SLIDE}: Reference-free evaluation for machine translation using a sliding document window.
\newblock In Kevin Duh, Helena Gomez, and Steven Bethard (eds.), \emph{Proceedings of the 2024 Conference of the North American Chapter of the Association for Computational Linguistics: Human Language Technologies (Volume 2: Short Papers)}, pp.\  205--211, Mexico City, Mexico, June 2024. Association for Computational Linguistics.
\newblock \doi{10.18653/v1/2024.naacl-short.18}.
\newblock URL \url{https://aclanthology.org/2024.naacl-short.18/}.

\bibitem[Rei et~al.(2020)Rei, Stewart, Farinha, and Lavie]{rei-etal-2020-comet}
Ricardo Rei, Craig Stewart, Ana~C Farinha, and Alon Lavie.
\newblock {COMET}: A neural framework for {MT} evaluation.
\newblock In Bonnie Webber, Trevor Cohn, Yulan He, and Yang Liu (eds.), \emph{Proceedings of the 2020 Conference on Empirical Methods in Natural Language Processing (EMNLP)}, pp.\  2685--2702, Online, November 2020. Association for Computational Linguistics.
\newblock \doi{10.18653/v1/2020.emnlp-main.213}.
\newblock URL \url{https://aclanthology.org/2020.emnlp-main.213/}.

\bibitem[Remy et~al.(2023)Remy, Delobelle, Berendt, Demuynck, and Demeester]{tiktotok23}
François Remy, Pieter Delobelle, Bettina Berendt, Kris Demuynck, and Thomas Demeester.
\newblock Tik-to-tok: Translating language models one token at a time: An embedding initialization strategy for efficient language adaptation, 2023.
\newblock URL \url{https://arxiv.org/abs/2310.03477}.

\bibitem[Schwenk et~al.(2021{\natexlab{a}})Schwenk, Chaudhary, Sun, Gong, and Guzm{\'a}n]{schwenk-etal-2021-wikimatrix}
Holger Schwenk, Vishrav Chaudhary, Shuo Sun, Hongyu Gong, and Francisco Guzm{\'a}n.
\newblock {W}iki{M}atrix: Mining 135{M} parallel sentences in 1620 language pairs from {W}ikipedia.
\newblock In Paola Merlo, Jorg Tiedemann, and Reut Tsarfaty (eds.), \emph{Proceedings of the 16th Conference of the European Chapter of the Association for Computational Linguistics: Main Volume}, pp.\  1351--1361, Online, April 2021{\natexlab{a}}. Association for Computational Linguistics.
\newblock \doi{10.18653/v1/2021.eacl-main.115}.
\newblock URL \url{https://aclanthology.org/2021.eacl-main.115/}.

\bibitem[Schwenk et~al.(2021{\natexlab{b}})Schwenk, Wenzek, Edunov, Grave, Joulin, and Fan]{schwenk-etal-2021-ccmatrix}
Holger Schwenk, Guillaume Wenzek, Sergey Edunov, Edouard Grave, Armand Joulin, and Angela Fan.
\newblock {CCM}atrix: Mining billions of high-quality parallel sentences on the web.
\newblock In Chengqing Zong, Fei Xia, Wenjie Li, and Roberto Navigli (eds.), \emph{Proceedings of the 59th Annual Meeting of the Association for Computational Linguistics and the 11th International Joint Conference on Natural Language Processing (Volume 1: Long Papers)}, pp.\  6490--6500, Online, August 2021{\natexlab{b}}. Association for Computational Linguistics.
\newblock \doi{10.18653/v1/2021.acl-long.507}.
\newblock URL \url{https://aclanthology.org/2021.acl-long.507}.

\bibitem[Sellam et~al.(2020)Sellam, Das, and Parikh]{sellam-etal-2020-bleurt}
Thibault Sellam, Dipanjan Das, and Ankur Parikh.
\newblock {BLEURT}: Learning robust metrics for text generation.
\newblock In Dan Jurafsky, Joyce Chai, Natalie Schluter, and Joel Tetreault (eds.), \emph{Proceedings of the 58th Annual Meeting of the Association for Computational Linguistics}, pp.\  7881--7892, Online, July 2020. Association for Computational Linguistics.
\newblock \doi{10.18653/v1/2020.acl-main.704}.
\newblock URL \url{https://aclanthology.org/2020.acl-main.704/}.

\bibitem[Sloto et~al.(2023)Sloto, Thompson, Khayrallah, Domhan, Gowda, and Koehn]{sloto-etal-2023-findings}
Steve Sloto, Brian Thompson, Huda Khayrallah, Tobias Domhan, Thamme Gowda, and Philipp Koehn.
\newblock Findings of the {WMT} 2023 shared task on parallel data curation.
\newblock In Philipp Koehn, Barry Haddow, Tom Kocmi, and Christof Monz (eds.), \emph{Proceedings of the Eighth Conference on Machine Translation}, pp.\  95--102, Singapore, December 2023. Association for Computational Linguistics.
\newblock \doi{10.18653/v1/2023.wmt-1.5}.
\newblock URL \url{https://aclanthology.org/2023.wmt-1.5/}.

\bibitem[Thompson \& Koehn(2020)Thompson and Koehn]{thompson-koehn-2020-exploiting}
Brian Thompson and Philipp Koehn.
\newblock Exploiting sentence order in document alignment.
\newblock In Bonnie Webber, Trevor Cohn, Yulan He, and Yang Liu (eds.), \emph{Proceedings of the 2020 Conference on Empirical Methods in Natural Language Processing (EMNLP)}, pp.\  5997--6007, Online, November 2020. Association for Computational Linguistics.
\newblock \doi{10.18653/v1/2020.emnlp-main.483}.
\newblock URL \url{https://aclanthology.org/2020.emnlp-main.483/}.

\bibitem[Thompson \& Post(2020{\natexlab{a}})Thompson and Post]{thompson-post-2020-automatic}
Brian Thompson and Matt Post.
\newblock Automatic machine translation evaluation in many languages via zero-shot paraphrasing.
\newblock In Bonnie Webber, Trevor Cohn, Yulan He, and Yang Liu (eds.), \emph{Proceedings of the 2020 Conference on Empirical Methods in Natural Language Processing (EMNLP)}, pp.\  90--121, Online, November 2020{\natexlab{a}}. Association for Computational Linguistics.
\newblock \doi{10.18653/v1/2020.emnlp-main.8}.
\newblock URL \url{https://aclanthology.org/2020.emnlp-main.8/}.

\bibitem[Thompson \& Post(2020{\natexlab{b}})Thompson and Post]{thompson-post-2020-paraphrase}
Brian Thompson and Matt Post.
\newblock Paraphrase generation as zero-shot multilingual translation: Disentangling semantic similarity from lexical and syntactic diversity.
\newblock In Lo{\"i}c Barrault, Ond{\v{r}}ej Bojar, Fethi Bougares, Rajen Chatterjee, Marta~R. Costa-juss{\`a}, Christian Federmann, Mark Fishel, Alexander Fraser, Yvette Graham, Paco Guzman, Barry Haddow, Matthias Huck, Antonio~Jimeno Yepes, Philipp Koehn, Andr{\'e} Martins, Makoto Morishita, Christof Monz, Masaaki Nagata, Toshiaki Nakazawa, and Matteo Negri (eds.), \emph{Proceedings of the Fifth Conference on Machine Translation}, pp.\  561--570, Online, November 2020{\natexlab{b}}. Association for Computational Linguistics.
\newblock URL \url{https://aclanthology.org/2020.wmt-1.67/}.

\bibitem[Thompson et~al.(2018)Thompson, Khayrallah, Anastasopoulos, McCarthy, Duh, Marvin, McNamee, Gwinnup, Anderson, and Koehn]{thompson-etal-2018-freezing}
Brian Thompson, Huda Khayrallah, Antonios Anastasopoulos, Arya~D. McCarthy, Kevin Duh, Rebecca Marvin, Paul McNamee, Jeremy Gwinnup, Tim Anderson, and Philipp Koehn.
\newblock Freezing subnetworks to analyze domain adaptation in neural machine translation.
\newblock In Ond{\v{r}}ej Bojar, Rajen Chatterjee, Christian Federmann, Mark Fishel, Yvette Graham, Barry Haddow, Matthias Huck, Antonio~Jimeno Yepes, Philipp Koehn, Christof Monz, Matteo Negri, Aur{\'e}lie N{\'e}v{\'e}ol, Mariana Neves, Matt Post, Lucia Specia, Marco Turchi, and Karin Verspoor (eds.), \emph{Proceedings of the Third Conference on Machine Translation: Research Papers}, pp.\  124--132, Brussels, Belgium, October 2018. Association for Computational Linguistics.
\newblock \doi{10.18653/v1/W18-6313}.
\newblock URL \url{https://aclanthology.org/W18-6313/}.

\bibitem[Thompson et~al.(2019{\natexlab{a}})Thompson, Gwinnup, Khayrallah, Duh, and Koehn]{thompson-etal-2019-overcoming}
Brian Thompson, Jeremy Gwinnup, Huda Khayrallah, Kevin Duh, and Philipp Koehn.
\newblock Overcoming catastrophic forgetting during domain adaptation of neural machine translation.
\newblock In Jill Burstein, Christy Doran, and Thamar Solorio (eds.), \emph{Proceedings of the 2019 Conference of the North {A}merican Chapter of the Association for Computational Linguistics: Human Language Technologies, Volume 1 (Long and Short Papers)}, pp.\  2062--2068, Minneapolis, Minnesota, June 2019{\natexlab{a}}. Association for Computational Linguistics.
\newblock \doi{10.18653/v1/N19-1209}.
\newblock URL \url{https://aclanthology.org/N19-1209/}.

\bibitem[Thompson et~al.(2019{\natexlab{b}})Thompson, Knowles, Zhang, Khayrallah, Duh, and Koehn]{thompson-etal-2019-hablex}
Brian Thompson, Rebecca Knowles, Xuan Zhang, Huda Khayrallah, Kevin Duh, and Philipp Koehn.
\newblock {HABL}ex: Human annotated bilingual lexicons for experiments in machine translation.
\newblock In Kentaro Inui, Jing Jiang, Vincent Ng, and Xiaojun Wan (eds.), \emph{Proceedings of the 2019 Conference on Empirical Methods in Natural Language Processing and the 9th International Joint Conference on Natural Language Processing (EMNLP-IJCNLP)}, pp.\  1382--1387, Hong Kong, China, November 2019{\natexlab{b}}. Association for Computational Linguistics.
\newblock \doi{10.18653/v1/D19-1142}.
\newblock URL \url{https://aclanthology.org/D19-1142/}.

\bibitem[Thompson et~al.(2024{\natexlab{a}})Thompson, Dhaliwal, Frisch, Domhan, and Federico]{thompson-etal-2024-shocking}
Brian Thompson, Mehak Dhaliwal, Peter Frisch, Tobias Domhan, and Marcello Federico.
\newblock A shocking amount of the web is machine translated: Insights from multi-way parallelism.
\newblock In Lun-Wei Ku, Andre Martins, and Vivek Srikumar (eds.), \emph{Findings of the Association for Computational Linguistics: ACL 2024}, pp.\  1763--1775, Bangkok, Thailand, August 2024{\natexlab{a}}. Association for Computational Linguistics.
\newblock \doi{10.18653/v1/2024.findings-acl.103}.
\newblock URL \url{https://aclanthology.org/2024.findings-acl.103/}.

\bibitem[Thompson et~al.(2024{\natexlab{b}})Thompson, Mathur, Deutsch, and Khayrallah]{thompson-etal-2024-improving}
Brian Thompson, Nitika Mathur, Daniel Deutsch, and Huda Khayrallah.
\newblock Improving statistical significance in human evaluation of automatic metrics via soft pairwise accuracy.
\newblock In Barry Haddow, Tom Kocmi, Philipp Koehn, and Christof Monz (eds.), \emph{Proceedings of the Ninth Conference on Machine Translation}, pp.\  1222--1234, Miami, Florida, USA, November 2024{\natexlab{b}}. Association for Computational Linguistics.
\newblock \doi{10.18653/v1/2024.wmt-1.118}.
\newblock URL \url{https://aclanthology.org/2024.wmt-1.118/}.

\bibitem[Touvron et~al.(2023)Touvron, Martin, Stone, Albert, Almahairi, Babaei, Bashlykov, Batra, Bhargava, Bhosale, Bikel, Blecher, Ferrer, Chen, Cucurull, Esiobu, Fernandes, Fu, Fu, Fuller, Gao, Goswami, Goyal, Hartshorn, Hosseini, Hou, Inan, Kardas, Kerkez, Khabsa, Kloumann, Korenev, Koura, Lachaux, Lavril, Lee, Liskovich, Lu, Mao, Martinet, Mihaylov, Mishra, Molybog, Nie, Poulton, Reizenstein, Rungta, Saladi, Schelten, Silva, Smith, Subramanian, Tan, Tang, Taylor, Williams, Kuan, Xu, Yan, Zarov, Zhang, Fan, Kambadur, Narang, Rodriguez, Stojnic, Edunov, and Scialom]{touvron2023llama}
Hugo Touvron, Louis Martin, Kevin Stone, Peter Albert, Amjad Almahairi, Yasmine Babaei, Nikolay Bashlykov, Soumya Batra, Prajjwal Bhargava, Shruti Bhosale, Dan Bikel, Lukas Blecher, Cristian~Canton Ferrer, Moya Chen, Guillem Cucurull, David Esiobu, Jude Fernandes, Jeremy Fu, Wenyin Fu, Brian Fuller, Cynthia Gao, Vedanuj Goswami, Naman Goyal, Anthony Hartshorn, Saghar Hosseini, Rui Hou, Hakan Inan, Marcin Kardas, Viktor Kerkez, Madian Khabsa, Isabel Kloumann, Artem Korenev, Punit~Singh Koura, Marie-Anne Lachaux, Thibaut Lavril, Jenya Lee, Diana Liskovich, Yinghai Lu, Yuning Mao, Xavier Martinet, Todor Mihaylov, Pushkar Mishra, Igor Molybog, Yixin Nie, Andrew Poulton, Jeremy Reizenstein, Rashi Rungta, Kalyan Saladi, Alan Schelten, Ruan Silva, Eric~Michael Smith, Ranjan Subramanian, Xiaoqing~Ellen Tan, Binh Tang, Ross Taylor, Adina Williams, Jian~Xiang Kuan, Puxin Xu, Zheng Yan, Iliyan Zarov, Yuchen Zhang, Angela Fan, Melanie Kambadur, Sharan Narang, Aurelien Rodriguez, Robert Stojnic, Sergey Edunov, and Thomas
  Scialom.
\newblock Llama 2: Open foundation and fine-tuned chat models, 2023.

\bibitem[Tran(2020)]{ramen20}
Ke~Tran.
\newblock From english to foreign languages: Transferring pre-trained language models, 2020.
\newblock URL \url{https://arxiv.org/abs/2002.07306}.

\bibitem[{\"U}st{\"u}n et~al.(2024){\"U}st{\"u}n, Aryabumi, Yong, Ko, D{'}souza, Onilude, Bhandari, Singh, Ooi, Kayid, Vargus, Blunsom, Longpre, Muennighoff, Fadaee, Kreutzer, and Hooker]{ustun-etal-2024-aya}
Ahmet {\"U}st{\"u}n, Viraat Aryabumi, Zheng Yong, Wei-Yin Ko, Daniel D{'}souza, Gbemileke Onilude, Neel Bhandari, Shivalika Singh, Hui-Lee Ooi, Amr Kayid, Freddie Vargus, Phil Blunsom, Shayne Longpre, Niklas Muennighoff, Marzieh Fadaee, Julia Kreutzer, and Sara Hooker.
\newblock Aya model: An instruction finetuned open-access multilingual language model.
\newblock In Lun-Wei Ku, Andre Martins, and Vivek Srikumar (eds.), \emph{Proceedings of the 62nd Annual Meeting of the Association for Computational Linguistics (Volume 1: Long Papers)}, pp.\  15894--15939, Bangkok, Thailand, August 2024. Association for Computational Linguistics.
\newblock URL \url{https://aclanthology.org/2024.acl-long.845}.

\bibitem[Vernikos et~al.(2022)Vernikos, Thompson, Mathur, and Federico]{vernikos-etal-2022-embarrassingly}
Giorgos Vernikos, Brian Thompson, Prashant Mathur, and Marcello Federico.
\newblock Embarrassingly easy document-level {MT} metrics: How to convert any pretrained metric into a document-level metric.
\newblock In \emph{Proceedings of the Seventh Conference on Machine Translation (WMT)}, pp.\  118--128, Abu Dhabi, United Arab Emirates (Hybrid), December 2022. Association for Computational Linguistics.
\newblock URL \url{https://aclanthology.org/2022.wmt-1.6/}.

\bibitem[Wang et~al.(2020)Wang, K, Mayhew, and Roth]{wang-etal-2020-extending}
Zihan Wang, Karthikeyan K, Stephen Mayhew, and Dan Roth.
\newblock Extending multilingual {BERT} to low-resource languages.
\newblock In Trevor Cohn, Yulan He, and Yang Liu (eds.), \emph{Findings of the Association for Computational Linguistics: EMNLP 2020}, pp.\  2649--2656, Online, November 2020. Association for Computational Linguistics.
\newblock \doi{10.18653/v1/2020.findings-emnlp.240}.
\newblock URL \url{https://aclanthology.org/2020.findings-emnlp.240}.

\bibitem[Wuebker et~al.(2018)Wuebker, Simianer, and DeNero]{wuebker-etal-2018-compact}
Joern Wuebker, Patrick Simianer, and John DeNero.
\newblock Compact personalized models for neural machine translation.
\newblock In Ellen Riloff, David Chiang, Julia Hockenmaier, and Jun{'}ichi Tsujii (eds.), \emph{Proceedings of the 2018 Conference on Empirical Methods in Natural Language Processing}, pp.\  881--886, Brussels, Belgium, October-November 2018. Association for Computational Linguistics.
\newblock \doi{10.18653/v1/D18-1104}.
\newblock URL \url{https://aclanthology.org/D18-1104/}.

\bibitem[Xu et~al.(2024)Xu, Kim, Sharaf, and Awadalla]{alma24}
Haoran Xu, Young~Jin Kim, Amr Sharaf, and Hany~Hassan Awadalla.
\newblock A paradigm shift in machine translation: Boosting translation performance of large language models.
\newblock In \emph{The Twelfth International Conference on Learning Representations}, 2024.
\newblock URL \url{https://openreview.net/forum?id=farT6XXntP}.

\bibitem[Yamaguchi et~al.(2024)Yamaguchi, Villavicencio, and Aletras]{yamaguchi2024empirical}
Atsuki Yamaguchi, Aline Villavicencio, and Nikolaos Aletras.
\newblock An empirical study on cross-lingual vocabulary adaptation for efficient language model inference, 2024.
\newblock URL \url{https://arxiv.org/abs/2402.10712}.

\bibitem[Zhang et~al.(2019)Zhang, Kishore, Wu, Weinberger, and Artzi]{bertscorepaper}
Tianyi Zhang, Varsha Kishore, Felix Wu, Kilian~Q. Weinberger, and Yoav Artzi.
\newblock Bertscore: Evaluating text generation with {BERT}.
\newblock \emph{CoRR}, abs/1904.09675, 2019.
\newblock URL \url{http://arxiv.org/abs/1904.09675}.

\bibitem[Zouhar et~al.(2024)Zouhar, Ding, Currey, Badeka, Wang, and Thompson]{zouhar-etal-2024-fine}
Vil{\'e}m Zouhar, Shuoyang Ding, Anna Currey, Tatyana Badeka, Jenyuan Wang, and Brian Thompson.
\newblock Fine-tuned machine translation metrics struggle in unseen domains.
\newblock In \emph{Proceedings of the 62nd Annual Meeting of the Association for Computational Linguistics (Volume 2: Short Papers)}, pp.\  488--500, Bangkok, Thailand, August 2024. Association for Computational Linguistics.
\newblock \doi{10.18653/v1/2024.acl-short.45}.
\newblock URL \url{https://aclanthology.org/2024.acl-short.45/}.

\end{thebibliography}
